\documentclass[conf]{IEEEtran}

\usepackage[shortcuts,acronym]{glossaries}
\newacronym{ABCD}{ABCD}{Apple Ball Cat Dog}
\usepackage{float}
\usepackage{graphicx}
\usepackage{caption}
\usepackage{subcaption}
\usepackage{tabularx}
\usepackage{lipsum}
\usepackage{balance}
\usepackage{adjustbox}
\usepackage{tikz}
\usepackage{booktabs}
\usepackage[ruled,vlined]{algorithm2e}

\usepackage{amsmath,amssymb,amsfonts}
\usepackage{algorithmic}
\usepackage{graphicx}
\usepackage{textcomp}
\usepackage{changepage} 
\usepackage{pgfplots}
\pgfplotsset{compat=1.17}
\usepackage{verbatim}
\def\BibTeX{{\rm B\kern-.05em{\sc i\kern-.025em b}\kern-.08em
    T\kern-.1667em\lower.7ex\hbox{E}\kern-.125emX}}
\usepackage[numbers,sort&compress]{natbib}
\usepackage{hyperref}
\hypersetup{
colorlinks   = true,
citecolor    = blue,
urlcolor    =  blue
}
\usepackage{url}
\usepackage{multirow}
\usepackage{hhline}
\usepackage{tabularx}
\newcolumntype{Y}{>{\centering\arraybackslash}X}
\newcolumntype{Z}{>{\raggedleft\arraybackslash}X}

\newcommand{\footURL}[1]{\footnote{\url{#1}}}

\makeatletter
\newcommand\footnoteref[1]{\protected@xdef\@thefnmark{\ref{#1}}\@footnotemark}
\makeatother

\usepackage{chapterbib}

\usepackage{url} 
 
\usepackage{breakurl}

\begin{document}

\title{Instruct-DeBERTa: A Hybrid Approach for Aspect-based Sentiment Analysis on Textual Reviews}

\DeclareRobustCommand*{\IEEEauthorrefmarkNum}[1]{%
  \raisebox{0pt}[0pt][0pt]{\textsuperscript{\footnotesize #1}}%
}

\author{
\IEEEauthorblockN{Dineth Jayakody\IEEEauthorrefmarkNum{1}\IEEEauthorrefmark{1},
A V A Malkith\IEEEauthorrefmarkNum{1}\IEEEauthorrefmark{1},
Koshila Isuranda\IEEEauthorrefmarkNum{1}\IEEEauthorrefmark{1},
Vishal Thenuwara\IEEEauthorrefmarkNum{2}\IEEEauthorrefmark{2},
Nisansa de Silva\IEEEauthorrefmarkNum{2}\IEEEauthorrefmark{2},\\
Sachintha Rajith Ponnamperuma\IEEEauthorrefmarkNum{3}\IEEEauthorrefmark{3},
G G N Sandamali\IEEEauthorrefmarkNum{1}\IEEEauthorrefmark{4},
K L K Sudheera\IEEEauthorrefmarkNum{1}\IEEEauthorrefmark{4}
}

\IEEEauthorblockA{\IEEEauthorrefmarkNum{1}\textit{Department of Electrical and Information Engineering}, \textit{University of Ruhuna}\\
\IEEEauthorrefmark{1}\texttt{\{jayakody\_ds\_e21,isuranda\_mak\_e21,malkith\_ava\_e21\}@engug.ruh.ac.lk}}\\
\IEEEauthorrefmark{4}\texttt{\{nadeesha,kushan\}@eie.ruh.ac.lk}\\
\IEEEauthorblockA{\IEEEauthorrefmarkNum{2}\textit{Department of Computer Science \& Engineering}, \textit{University of Moratuwa}\\
\IEEEauthorrefmark{2}\texttt{\{vishal.23,NisansaDdS\}@cse.mrt.ac.lk}}\\
\IEEEauthorblockA{\IEEEauthorrefmarkNum{3}\textit{Emojot Inc.}\\
\IEEEauthorrefmark{3}\texttt{sachintha@emojot.com}}
}


\maketitle

\begin{abstract}

Aspect-based Sentiment Analysis (ABSA) is a critical task in Natural Language Processing (NLP) that focuses on extracting sentiments related to specific aspects within a text, offering deep insights into customer opinions. Traditional sentiment analysis methods, while useful for determining overall sentiment, often miss the implicit opinions about particular product or service features. This paper presents a comprehensive review of the evolution of ABSA methodologies, from lexicon-based approaches to machine learning and deep learning techniques. We emphasize the recent advancements in Transformer-based models, particularly Bidirectional Encoder Representations from Transformers (BERT) and its variants, which have set new benchmarks in ABSA tasks.
We focused on finetuning Llama and Mistral models, building hybrid models using the SetFit framework, and developing our own model by exploiting the strengths of state-of-the-art (SOTA) Transformer-based models for aspect term extraction (ATE) and aspect sentiment classification (ASC).
Our hybrid model \texttt{Instruct - DeBERTa} uses SOTA \texttt{InstructABSA} for aspect extraction and \texttt{DeBERTa-V3-baseabsa-V1} for aspect sentiment classification. We utilize datasets from different domains to evaluate our model's performance. Our experiments indicate that the proposed hybrid model significantly improves the accuracy and reliability of sentiment analysis across all experimented domains. As per our findings, our hybrid model \texttt{Instruct - DeBERTa} is the best-performing model for the joint task of ATE and ASC for both SemEval restaurant 2014 and SemEval laptop 2014 datasets separately.
By addressing the limitations of existing methodologies, our approach provides a robust solution for understanding detailed consumer feedback, thus offering valuable insights for businesses aiming to enhance customer satisfaction and product development.

\end{abstract}


\IEEEpeerreviewmaketitle

\section{Introduction}

Aspect-Based Sentiment Analysis (ABSA) has become an essential technique in Natural Language Processing (NLP) for extracting fine-grained opinions from textual data. It focuses on identifying sentiment towards specific aspects within a text, providing a detailed understanding of customer feedback and reviews. Traditional sentiment analysis techniques, while effective at determining overall sentiment, often fail to capture the nuanced opinions that consumers express about particular features or attributes of a product or service.

Over the years, ABSA methodologies have evolved significantly. Early approaches primarily relied on lexicon-based methods, which used predefined dictionaries of sentiment-laden words to infer polarity. These methods, however, struggled with context and ambiguity. The advent of machine learning introduced more sophisticated techniques, including supervised learning models that could be trained on annotated datasets. Despite their advancements, these models required substantial manual effort for feature extraction and were often domain-specific.

The breakthrough in deep learning, particularly with the development of recurrent neural networks (RNNs), Long Short-Term Memory networks (LSTMs), and Convolutional Neural Networks (CNNs), marked a significant improvement in sentiment analysis. These models could automatically learn features from text, capturing context and sequential dependencies more effectively than traditional methods. LSTM and CNN-based models became popular for their ability to handle long-range dependencies and local features, respectively. However, these models still had limitations, especially in understanding long-term dependencies and complex syntactic structures.

The introduction of Transformer architectures, especially \texttt{BERT}, revolutionized the field by leveraging attention mechanisms to capture contextual relationships in both directions of a sentence. \texttt{BERT} and its variants, such as \texttt{RoBERTa} and \texttt{DeBERTa}, have set new benchmarks in various NLP tasks, including ABSA. These models have demonstrated superior performance in aspect extraction and sentiment classification tasks due to their ability to understand complex language patterns and relationships.

In our study, we focus on developing a hybrid model that leverages the strengths of the latest Transformer-based models for ABSA. We aim to address the limitations of existing approaches by combining aspect extraction and sentiment classification into a unified framework. Our approach utilizes datasets from the hospitality domain, including SemEval 2014 (Res-14), 2015 (Res-15), and 2016 (Res-16) restaurant reviews, and extends to the laptop domain with the SemEval 2014 laptop dataset (Lap-14). By evaluating these models on imbalanced datasets using the F1 metric, we ensure a balanced and comprehensive assessment of performance.

Through our literature review, we have identified key methodologies and their respective accuracies, guiding the design of our hybrid model. We focus on models that excel in aspect term extraction (ATE) and aspect sentiment classification (ASC), aiming to develop a model that builds on the successes of past methodologies while innovating in areas where existing methods may fall short. Our goal is to enhance the accuracy and reliability of sentiment analysis in our application domain, ultimately providing a robust solution for understanding consumer feedback.

\section{Literature Review}

Our literature review systematically investigates various models that have demonstrated efficacy in ATE and ASC. 
\autoref{tab:tablenewlitreview} provides a comprehensive summary of the methodologies employed for these tasks over the years. Notably, this table exclusively considers models that do not incorporate LLMs. \autoref{table:f1_scores_new} lists down models that perform the joint task which is the ATE and ASC tasks together by a single model. The model is only fed in with the relevant sentences or the reviews. Then the model identifies the aspects by itself and classifies the polarities to the aspects that have been identified. Then the F1 score of the whole process is reported. For a sentence $S_i$ the ATE, ASC, and the joint task can be visualized as below.

\noindent
$S_i$: The \textcolor{blue}{\underline{price}} was \textcolor{red}{\underline{high}}, but the \textcolor{blue}{\underline{restaurant}} was \textcolor{red}{\underline{breathtaking}}.

\begin{table}[ht]
\centering
\caption{Overview of tasks for ABSA with the sample sentence $S_i$}
\label{table:subtasksNew}
\resizebox{\linewidth}{!}{
\renewcommand{\arraystretch}{2.0} 
\fontsize{16}{18}\selectfont 
\begin{tabular}{|c|c|c|}
\hline
\textbf{Task} & \textbf{Input} & \textbf{Output} \\ \hline
\textbf{Aspect Term Extraction (ATE)} & $S_i$ & price, restaurant \\ \hline
\textbf{Aspect Sentiment Classification (ASC)} & $S_i + price, S_i + restaurant$ & Negative, Positive \\ \hline
\textbf{Joint Task (ATE + ASC)} & $S_i$ & (price, Negative), (restaurant, Positive) \\ \hline
\end{tabular}
}
\end{table}

\subsection{Models for a Single Task (ATE or ASC)}
\subsubsection{\textbf{LSTM Based Models}}

\vspace{1pt}

Standard RNNs suffer from significant limitations, primarily the vanishing gradient and exploding gradient problems. To address these limitations, the LSTM network was developed by~\citet{hochreiter1997long}. To effectively utilize aspect information, ~\citet{wang2016attention} proposed a model called LSTM with Aspect Embedding (\texttt{AE-LSTM}). 
However, to further leverage aspect information, ~\citet{wang2016attention} developed an enhanced model called Attention-based LSTM with Aspect Embedding (\texttt{ATAE-LSTM}). 

~\citet{li2018transformation} proposed Target-Specific Transformation Networks (\texttt{TNet}), a new architecture designed to improve target sentiment classification by effectively handling multiple targets and extracting relevant features without introducing noise. \texttt{TNet} introduces a novel Target-Specific Transformation (TST)~\cite{li2018transformation} component for generating target-specific word representations. 
The models \texttt{LSTM-FC-CNN-LF} and \texttt{LSTM-FC-CNN-AS} were built by ~\citet{li2018transformation} that incorporate a fully connected layer and context-preserving mechanisms. These models performed better with F1 scores of 70.60\% and 70.23\% for \texttt{LSTM-FC-CNN-LF}, 70.72\% and 70.06\% for \texttt{LSTM-FC-CNN-AS} for Lap-14 and Res-14 respectively.

\subsubsection{\textbf{GloVe Based Models}}
\vspace{1pt}
\texttt{GloVe} (Global Vectors for Word Representation) is an algorithm that generates word embeddings by aggregating word co-occurrence statistics from a corpus. It is important because it captures both local and global statistical information of words, enhancing the performance of natural language processing tasks. 
\texttt{ASGCN} introduced by~\citet{zhang-etal-2019-aspect} proposed a novel aspect-specific sentiment classification framework while \texttt{DualGCN} by~\citet{li2021dual} proposed a dual graph convolutional network model that considers the complementarity of syntax structures and semantic correlations simultaneously.~\citet{zhong2023KGAN} proposed a new model named \texttt{KGAN} which uses a knowledge graph augmented network, which aims to effectively incorporate external knowledge with explicitly syntactic and contextual information.
While all these models individually had their own importance, merging them with Glove as an embedding method, researchers were able to capture both local and global information to achieve higher F1 Scores like 84.46\% with Res-14. However, \texttt{UIKA} (Unified Instance and Knowledge Alignment Pretraining) by \citet{liu2023unified}, which introduces a unified alignment pretraining framework into the vanilla pretrain-finetune pipeline, incorporates both instance and knowledge level alignments. It reported a higher F1 score of 85.53\% with \texttt{KGAN} for Res-14. Furthermore, \texttt{KGAN+UIKA} achieved higher F1 scores with \texttt{BERT} compared to \texttt{GloVE}.

\subsubsection{\textbf{BERT Based Models}}

\vspace{1pt}

The original \texttt{BERT} model was introduced by~\citet{devlin2018bert}. \texttt{BERT}’s training process involves two main steps: pre-training and fine-tuning. \texttt{BERT-DK}, introduced by~\citet{zhao2020bertdk}, integrates domain-specific knowledge to improve ABSA performance. By incorporating domain-specific information, \texttt{BERT-DK} achieved F1 scores of 77.02\% and 83.55\% for aspect extraction on the Res-14 and Lap-14 datasets respectively. 
Similarly, \texttt{BERT-SPC}, developed by~\citet{song2019attentional}, employs a Sentence Pair Classification framework to better understand the context of aspect-specific sentences. 

Innovative approaches such as \texttt{BERT-MRC}, proposed by \citet{bert_mrc_2020}, frame ABSA tasks as machine reading comprehension problems while ~\citet{xu2019bert} introduced \texttt{BERT-PT} which involves pre-training \texttt{BERT} on domain-specific data followed by fine-tuning. 
\texttt{BAT} which stands for \texttt{BERT} with Adversarial Training, introduced by ~\citet{karimi2021adversarial}, enhances ABSA by generating adversarial examples during training. 


Cutting-edge models like \texttt{RGAT-BERT}, \texttt{DualGCN-BERT}, \texttt{TF-BERT}, and \texttt{dotGCN-BERT} have further improved ABSA performance. \texttt{RGAT-BERT}, proposed by~\citet{bai2020investigating}, uses relational graph attention networks to improve aspect extraction and sentiment classification abilities. 
In addition to that \texttt{DualGCN-BERT} introduced by~\citet{li2021dual}, uses dual graph convolutional networks to handle both aspect extraction and sentiment classification. 
\texttt{TF-BERT}, developed by~\citet{zhang2023span}, uses task-specific fine-tuning strategies to improve ABSA performance. 
In contrast to that \texttt{DotGCN-BERT}, proposed by ~\citet{chen2022discrete}, uses dot-product based graph convolutional networks to improve ABSA performance. 
Furthermore keeping another step forward \texttt{DualGCN and KGAN} have used \texttt{BERT} as the embedding metholody in order to achieve higher F1 scores. \autoref{tab:tablenewlitreview}  shows that \texttt{KGAN+UIKA} with \texttt{BERT} reported a F1-score of 92.89\%, which is the highest for the Res-16 dataset. But still we decided to move forward with \texttt{DeBERTa-V3-base-absa-v1} since it provides a higher F1 Score in all datasets compared to others.


\subsubsection{\textbf{RoBERTa Based Models}}
\vspace{1pt}

\texttt{RoBERTa}~\cite{liu2019roberta} is an advanced language model that builds upon the foundational work of \texttt{BERT}. The \texttt{SARL-RoBERTa} model, which was introduced by~\citet{wang2021eliminating} used span-based dependency modeling to align opinion candidates with aspects and used an adversarial learning strategy to reduce sentiment bias in aspect embeddings. Among the compared \texttt{RoBERTa} based models, \texttt{SARL-RoBERTa} performs the best, achieving a F1-score of 82.44\% and 82.97\% for Res-14 and Lap-14 respectively. 

However, models such as \texttt{ASGCN-RoBERTa}, \texttt{RGAT-RoBERTa}, \texttt{PWCN-RoBERTa}, and \texttt{RoBERTa+MLP} benefited by combining \texttt{RoBERTa} with various specialized architectures, as demonstrated by~\citet{dai2021does}. Strong performance is achieved by \texttt{ASGCN-RoBERTa}, which combines an aspect-specific graph convolutional network with dependency tree syntactic information. With a relational graph attention network integrated to collect relational information between words, \texttt{RGAT-RoBERTa} performs admirably. \texttt{RoBERTa+MLP} integrates a multi-layer perceptron with \texttt{RoBERTa}. It highlights the flexibility of combining \texttt{RoBERTa's} embeddings with simple classifiers. 

Task-oriented syntactic information is well captured by pure \texttt{RoBERTa} based models, particularly by the fine-tuned variations (\texttt{FT-RoBERTa}). Research conducted by~\citet{dai2021does} demonstrates that \texttt{FT-RoBERTa} achieves a 1.56\% improvement in the F1-score over standard \texttt{RoBERTa} induced trees, and performs better than parser-provided trees. 

\subsubsection{\textbf{DeBERTa Based Models}}
\vspace{1pt}

\texttt{DeBERTa}~\cite{he2020deberta} introduces a disentangled attention mechanism, which utilizes two separate vectors for each word to represent its content and position independently. 
Also, the model incorporates an enhanced mask decoder in its pre-training phase based on masked language modeling (MLM). 



Improving from the vanilla \texttt{DeBERTa} model a new model named \texttt{DeBERTaV3} was introduced by~\citet{he2021debertav3}. 
By further fine tuning the model to improve its performance \citet{DBLP:conf/cikm/0008ZL23} developed the \texttt{DeBERTaV3-base-absa-V1} model. This was trained using Lap-14, Res-14, Res-16, and six more datasets counting up to 30k+ ABSA examples. The accuracy of this model showed an improvement of 9.35\% and 10.87\% for the ASC task of Res-14 and Lap-14 datasets respectively compared to the original \texttt{DeBERTaV3} model.

In their independent investigations,~\citet{marcacini2021aspect} as well as ~\citet{yang2024modeling} explored the utilization of \texttt{DeBERTa} based models, introducing \texttt{ABSA-DeBERTa} and \texttt{LSA-X-DeBERTa}, respectively. \citet{marcacini2021aspect} explored disentangled learning as a method to improve \texttt{BERT-based} representations specifically for ABSA.
On the other hand, \citet{yang2024modeling} introduced a novel perspective in ASC by emphasizing the significance of aspect sentiment coherency. Their study revealed that neighbouring aspects usually share similar sentiments, which is known as ``aspect sentiment coherency." To address this, they proposed a local sentiment aggregation paradigm (\texttt{LSA}) to effectively model fine-grained sentiment coherency. In respect to that, the \texttt{LSA-X-DeBERTa} model introduced by \citet{yang2024modeling} achieved a F1-score of 87.02\% for Res-14 and 84.41\% for Lap-14 under the sentiment classification task.

\subsubsection{\textbf{Other Models}}

\vspace{1pt}
Concerning ASC and ATE, in particular, \texttt{LCF-ATEPC-CDM} proposed by~\citet{yang2021multi} and \texttt{InstructABSA} proposed by~\citet{scaria2023instructabsa} standout for their strong performances. \texttt{InstructABSA}, utilizing a novel instruction learning paradigm, showed exceptional abilities in obtaining pertinent aspects from the text, attaining F1 scores exceeding 92\% for aspect extraction on both the Res-14 and Lap-14 datasets. Concerning ATE, \texttt{LCF-ATEPC-CDM}, which also employs a local context focus technique, performs fairly well.
In sentiment polarity classification, \texttt{InstructABSA} also excels with F1 scores of 85.17\% on Res-14 and 81.56\% on Lap-14, outperforming many other models. 
The \texttt{LSAT} model proposed by~\citet{yang2021improving}, with its focus on aspect sentiment coherency through a local sentiment aggregation paradigm, shows impressive results, achieving a F1 score of 90.86\% on Res-14.
The efficacy of the \texttt{BART-ABSA} model in a comprehensive approach to ABSA has also been demonstrated by~\citet{yan2021unified}, which combines all ABSA subtasks into a single generative formulation. 
\subsection{Joint Task Models}

Here we talk about the models that perform the joint task which is the ATE and ASC tasks together by a single model. The model is only fed in with the relevant sentences or the reviews. Then the model identifies the aspects by itself and classifies the polarities to the aspects that have been identified. Then the F1 score of the whole process is reported.

\autoref{table:f1_scores_new} presents the leading joint task models identified through our research. The \texttt{InstructABSA} and \texttt{Grace} models, which were previously described as single task models capable of performing both ATE and ASC tasks separately, also excel in the joint task. These models report the highest F1 scores for the joint task, achieving over 75\% for both datasets, indicating their accuracy and robustness across different domains.

\texttt{RACL-BERT}, introduced by \citet{chen2020relation}, is a notable ABSA (Aspect-Based Sentiment Analysis) model that utilizes the \texttt{BERT-Large} model to address three subtasks simultaneously: identifying aspects, detecting sentiment words, and classifying overall sentiment. Through multitasking and relation propagation, RACL-BERT enhances sentiment analysis accuracy. Similarly, \texttt{SPAN}, introduced by \citet{hu2019open}, employs a novel approach by focusing on key opinion points rather than tagging each word. Both \texttt{RACL-BERT} and \texttt{SPAN} achieved reasonable F1 scores but were outperformed by \texttt{InstructABSA} and \texttt{GRACE}.

The \texttt{E2E-TBSA} model, proposed by \citet{li2019unified}, addresses both ATE and ASC tasks in a single step, using a collapsed approach that combines these tasks into a unified process. Similarly, \texttt{BERT-E2E-ABSA}, introduced by \citet{li2019exploiting}, is based on \texttt{BERT} models and follows the same principles. These models achieved F1 scores in the 60\%-70\% range but did not outperform the leading models.

\texttt{DOER}, introduced by \citet{luo2019doer}, uses a cross-shared RNN framework to generate aspect term-polarity pairs simultaneously. \texttt{IMN}, introduced by \citet{he2019interactive}, employs an interactive architecture with multi-task learning for end-to-end ABSA tasks, including aspect term and opinion term extraction as well as aspect-level sentiment classification.

\subsection{Selection Criteria: Dataset}
After a thorough review, the following criteria were established for dataset selection:

\begin{itemize}
    \item Relevance to BSA:
The datasets must be specifically designed for or widely used in aspect-based sentiment analysis ensuring granularity.
    \item Diversity of aspects and sentiments:
The selected datasets should cover a wide range of aspects and sentiments ensuring generalizability.
    \item Quality of annotations:
High-quality, manually annotated datasets are preferred to ensure the accuracy. 
    \item Availability and accessibility:
Publicly available datasets with accessibility are chosen to facilitate reproducibility.
\end{itemize}

Based on these criteria, the SemEval datasets from the years 2014, 2015, and 2016 were selected.

\begin{table*}[!htbp]
\caption{F1 scores of models evaluated on the SemEval 2014~\cite{pontiki-etal-2014-semeval} benchmark. Note$^*$: The F1-scores for the DeBERTa-v3-base-absa-v1 model were calculated by us separately.} 
\label{tab:tablenewlitreview}
\begin{adjustbox}{width=0.78\textwidth, center}
\scriptsize
\begin{tabular}{|l|l|c|c|c|c|c|c|c|c|}
\hline
\multicolumn{2}{|l|}{\multirow{3}{*}{\textbf{Model}}} & \multicolumn{8}{c|}{\textbf{F1 Score (\%)}} \\
\hhline{~~--------}
\multicolumn{2}{|l|}{} & \multicolumn{2}{c|}{\textbf{Res-14}} & \multicolumn{2}{c|}{\textbf{Lap-14}} & \multicolumn{2}{c|}{\textbf{Res-15}} & \multicolumn{2}{c|}{\textbf{Res-16}} \\
\hhline{~~--------}
\multicolumn{2}{|l|}{} & \textbf{ATE} & \textbf{ASC} & \textbf{ATE} & \textbf{ASC} & \textbf{ATE} & \textbf{ASC} & \textbf{ATE} & \textbf{ASC} \\
\hline
\multicolumn{2}{|l|}{\textbf{InstructABSA}~\cite{scaria2023instructabsa}} & \textbf{92.10} & 85.17 & \textbf{92.30} & 81.56 & \textbf{76.64} & 84.50 & \textbf{80.32} & 89.43  \\ 
\hline
\multicolumn{2}{|l|}{GRACE~\cite{luo2020grace}} & 85.45 &  \textbf{-\!-\!-} & 87.93 &  \textbf{-\!-\!-} &  \textbf{-\!-\!-} &  \textbf{-\!-\!-} &  \textbf{-\!-\!-} &  \textbf{-\!-\!-}  \\ 
\hline
\multicolumn{2}{|l|}{LCF-ATEPC-CDM~\cite{yang2021multi}} & 89.78 & 85.88 & 85.29 & 79.84 & \textbf{-\!-\!-}  & \textbf{-\!-\!-}  & \textbf{-\!-\!-}  & \textbf{-\!-\!-}  \\ 
\hline
\multicolumn{2}{|l|}{MCRF-SA~\cite{xu2020aspect}} &  \textbf{-\!-\!-} &73.78  & \textbf{-\!-\!-} &74.23 & \textbf{-\!-\!-} & 61.59 & \textbf{-\!-\!-} & 75.92  \\ 
\hline
\multicolumn{2}{|l|}{KaGRMN-DSG~\cite{xing2022understand}} & \textbf{-\!-\!-}  & 81.98 & \textbf{-\!-\!-} & 79.42 & \textbf{-\!-\!-} & \textbf{-\!-\!-}&\textbf{-\!-\!-} & \textbf{-\!-\!-}  \\ 
\hline
\multicolumn{2}{|l|}{MGAN ~\cite{li2019exploiting}} &\textbf{-\!-\!-}  & 71.48 &\textbf{-\!-\!-}  &71.42 & \textbf{-\!-\!-} & \textbf{-\!-\!-} & \textbf{-\!-\!-} & \textbf{-\!-\!-} \\ 
\hline
\multicolumn{2}{|l|}{IAN~\cite{ma2017interactive}} &  \textbf{-\!-\!-} &70.09  & \textbf{-\!-\!-} &67.38 & \textbf{-\!-\!-} & 52.65 & \textbf{-\!-\!-} & 55.21  \\ 
\hline
\multicolumn{2}{|l|}{BART-ABSA~\cite{yan2021unified}} & 87.07 & \textbf{-\!-\!-} & 83.52 & \textbf{-\!-\!-}& 75.48 & \textbf{-\!-\!-} & \textbf{-\!-\!-}  & \textbf{-\!-\!-} \\ 
\hline
\multicolumn{2}{|l|}{LSAT~\cite{yang2021improving}} & \textbf{-\!-\!-} & 90.86  & \textbf{-\!-\!-}  & 86.31 & \textbf{-\!-\!-}  & \textbf{-\!-\!-}  & \textbf{-\!-\!-}  & \textbf{-\!-\!-}  \\ 
\hline
\hline
\multicolumn{2}{|l|}{ASGCN-GloVe~\cite{zhang-etal-2019-aspect}} & \textbf{-\!-\!-} & 80.86 & \textbf{-\!-\!-}  & 75.55 & \textbf{-\!-\!-} &  79.34 & \textbf{-\!-\!-}  & 88.69 \\ 
\hline
\multicolumn{2}{|l|}{DualGCN-GloVe~\cite{li2021dual}} & \textbf{-\!-\!-} & 84.27  & \textbf{-\!-\!-} & 78.48 & \textbf{-\!-\!-} & 79.11 & \textbf{-\!-\!-} & 87.80 \\ 
\hline
\multicolumn{2}{|l|}{DualGCN-GloVe+UIKA~\cite{liu2023unified}} & \textbf{-\!-\!-} & 85.19  & \textbf{-\!-\!-} & 78.89 & \textbf{-\!-\!-} & 81.16 & \textbf{-\!-\!-} & 88.91\\ 
\hline
\multicolumn{2}{|l|}{KGAN-GloVe~\cite{zhong2023KGAN}} & \textbf{-\!-\!-} & 84.46  & \textbf{-\!-\!-} & 78.91 & \textbf{-\!-\!-} & 83.09 & \textbf{-\!-\!-} & 89.78 \\ 
\hline
\multicolumn{2}{|l|}{KGAN-GloVe+UIKA~\cite{liu2023unified}} & \textbf{-\!-\!-} & 85.53  & \textbf{-\!-\!-} & 79.31 & \textbf{-\!-\!-} & 83.89 & \textbf{-\!-\!-} & 90.92 \\ 
\hline
\hline
\multicolumn{2}{|l|}{DeBERTaV3~\cite{he2020deberta}~\cite{he2021debertav3}} & \textbf{-\!-\!-} & 83.06 & \textbf{-\!-\!-} & 79.45 & \textbf{-\!-\!-} & 73.76 & \textbf{-\!-\!-} & 73.59 \\ 
\hline
\multicolumn{2}{|l|}{ABSA-DeBERTa\cite{marcacini2021aspect}} & \textbf{-\!-\!-}  & 89.46  & \textbf{-\!-\!-}  & 82.76 & \textbf{-\!-\!-} & \textbf{-\!-\!-} & \textbf{-\!-\!-} & \textbf{-\!-\!-}  \\ 
\hline
\multicolumn{2}{|l|}{\textbf{DeBERTa-V3-base-absa-v1$^*$}~\cite{DBLP:conf/cikm/0008ZL23,YangZMT21} } & \textbf{-\!-\!-} & \textbf{90.94} & \textbf{-\!-\!-} & \textbf{90.32} & \textbf{-\!-\!-} & \textbf{89.55} & \textbf{-\!-\!-} & \textbf{84.91}\\ 
\hline
\multicolumn{2}{|l|}{LSA-X-DeBERTa~\cite{yang2024modeling}} & \textbf{-\!-\!-} & 87.02  & \textbf{-\!-\!-}  & 84.41 & \textbf{-\!-\!-} & 81.29 & \textbf{-\!-\!-} & 84.87  \\ 
\hline
\hline
\multicolumn{2}{|l|}{BERT~\cite{devlin2018bert}} & \textbf{-\!-\!-}  & 71.91  &79.28 & 71.94 & \textbf{-\!-\!-} & \textbf{-\!-\!-} & 74.10 & \textbf{-\!-\!-} \\ 
\hline
\multicolumn{2}{|l|}{BERT-DK~\cite{zhao2020bertdk}} &77.02  &75.45  &83.55  &73.72 & \textbf{-\!-\!-} & \textbf{-\!-\!-} & \textbf{-\!-\!-} & \textbf{-\!-\!-} \\ 
\hline
\multicolumn{2}{|l|}{BERT-SPC~\cite{song2019attentional}~\cite{karimi2020improving}} & \textbf{-\!-\!-} & 76.98 & \textbf{-\!-\!-} & 75.03 & \textbf{-\!-\!-} & \textbf{-\!-\!-} & \textbf{-\!-\!-} & \textbf{-\!-\!-}\\ 
\hline
\multicolumn{2}{|l|}{BERT-MRC~\cite{bert_mrc_2020}} &74.21  &74.97  &81.06  & 74.10 & \textbf{-\!-\!-} & \textbf{-\!-\!-}  & \textbf{-\!-\!-}  & \textbf{-\!-\!-}\\ 
\hline
\multicolumn{2}{|l|}{BERT-PT~\cite{xu2019bert}~\cite{karimi2021adversarial}} & \textbf{-\!-\!-} &76.96  &84.26 & 75.08 &  \textbf{-\!-\!-} &  \textbf{-\!-\!-} & 81.57 &  \textbf{-\!-\!-} \\ 
\hline
\multicolumn{2}{|l|}{BAT~\cite{karimi2021adversarial}} & \textbf{-\!-\!-} & 76.50 & 85.57 & 79.24 & \textbf{-\!-\!-} & \textbf{-\!-\!-} & 81.50 & \textbf{-\!-\!-} \\ 
\hline
\multicolumn{2}{|l|}{P-SUM~\cite{karimi2020improving}} & \textbf{-\!-\!-} & 79.68 & 85.94 & 76.81 &  \textbf{-\!-\!-} &  \textbf{-\!-\!-} &  81.99 & \textbf{-\!-\!-}\\ 
\hline
\multicolumn{2}{|l|}{H-SUM~\cite{karimi2020improving}} & \textbf{-\!-\!-} & 79.67 & 86.09 & 76.52 & \textbf{-\!-\!-} & \textbf{-\!-\!-} & 81.34 & \textbf{-\!-\!-} \\ 
\hline
\multicolumn{2}{|l|}{SK-GCN-BERT~\cite{zhou2020sk}} & \textbf{-\!-\!-} & 75.19 & \textbf{-\!-\!-}  & 75.57 &  \textbf{-\!-\!-} & 66.78 & \textbf{-\!-\!-} & 72.02 \\ 
\hline
\multicolumn{2}{|l|}{SDGCN-BERT~\cite{zhao2020modeling}} & \textbf{-\!-\!-} & 76.47  & \textbf{-\!-\!-} & 78.34 & \textbf{-\!-\!-} & \textbf{-\!-\!-} & \textbf{-\!-\!-} & \textbf{-\!-\!-} \\ 
\hline
\multicolumn{2}{|l|}{RGAT-BERT~\cite{bai2020investigating}} & \textbf{-\!-\!-} & 80.92 & \textbf{-\!-\!-}  & 78.20 & \textbf{-\!-\!-} &  66.18 & \textbf{-\!-\!-}  & 71.13 \\ 
\hline
\multicolumn{2}{|l|}{RGAT-BERT+UIKA~\cite{liu2023unified}} & \textbf{-\!-\!-} & 87.25 & \textbf{-\!-\!-}  & 82.03 & \textbf{-\!-\!-} & 86.40 & \textbf{-\!-\!-}  &  91.87 \\
\hline
\multicolumn{2}{|l|}{DGEDT-BERT~\cite{tang2020dependency}} & \textbf{-\!-\!-}  & 80.00 & \textbf{-\!-\!-} & 75.60 & \textbf{-\!-\!-} &  71.00 & \textbf{-\!-\!-} & 79.00  \\ 
\hline
\multicolumn{2}{|l|}{DualGCN-BERT~\cite{li2021dual}} & \textbf{-\!-\!-} & 81.16  & \textbf{-\!-\!-} & 78.10 & \textbf{-\!-\!-} & 84.25 & \textbf{-\!-\!-} & 89.22 \\ 
\hline
\multicolumn{2}{|l|}{DualGCN-BERT+UIKA~\cite{liu2023unified}} & \textbf{-\!-\!-} & 87.90  & \textbf{-\!-\!-} & 82.66 & \textbf{-\!-\!-} & 86.21 & \textbf{-\!-\!-} & 90.81 \\ 
\hline
\multicolumn{2}{|l|}{KGAN-BERT~\cite{zhong2023KGAN}} & \textbf{-\!-\!-} & 87.15  & \textbf{-\!-\!-} & 78.10 & \textbf{-\!-\!-} & 84.25 & \textbf{-\!-\!-} & 92.34 \\ 
\hline
\multicolumn{2}{|l|}{KGAN-BERT+UIKA~\cite{liu2023unified}} & \textbf{-\!-\!-} & 87.92  & \textbf{-\!-\!-} & 83.21 & \textbf{-\!-\!-} & 87.43 & \textbf{-\!-\!-} & 92.89 \\ 
\hline

\multicolumn{2}{|l|}{DualGCN-BERT~\cite{li2021dual}} & \textbf{-\!-\!-} & 81.16  & \textbf{-\!-\!-} & 78.10 & \textbf{-\!-\!-} & \textbf{-\!-\!-} & \textbf{-\!-\!-} & \textbf{-\!-\!-} \\ 
\hline
\multicolumn{2}{|l|}{BERT-ADA~\cite{rietzler2019adapt}} & \textbf{-\!-\!-} & 80.05 & \textbf{-\!-\!-} & 74.09 & \textbf{-\!-\!-} & \textbf{-\!-\!-} & \textbf{-\!-\!-} & \textbf{-\!-\!-} \\ 
\hline
\multicolumn{2}{|l|}{TF-BERT~\cite{zhang2023span}} & \textbf{-\!-\!-} & 81.15 & \textbf{-\!-\!-}  & 78.46 & \textbf{-\!-\!-} & \textbf{-\!-\!-}& \textbf{-\!-\!-} & \textbf{-\!-\!-} \\ 
\hline
\multicolumn{2}{|l|}{Dual-MRC~\cite{mao2021joint}} & \textbf{-\!-\!-} & 82.04 & \textbf{-\!-\!-} & 75.97 & \textbf{-\!-\!-} & \textbf{-\!-\!-} & \textbf{-\!-\!-} & \textbf{-\!-\!-} \\ 
\hline
\multicolumn{2}{|l|}{dotGCN-BERT~\cite{chen2022discrete}} & \textbf{-\!-\!-}  & 80.49  & \textbf{-\!-\!-} & 78.10 &\textbf{-\!-\!-} & \textbf{-\!-\!-} & \textbf{-\!-\!-}& \textbf{-\!-\!-}  \\ 
\hline
\multicolumn{2}{|l|}{DPL-BERT~\cite{zhang2022towards}} & \textbf{-\!-\!-} & 84.86 & \textbf{-\!-\!-}  & 78.58 &\textbf{-\!-\!-} & \textbf{-\!-\!-} & \textbf{-\!-\!-}& \textbf{-\!-\!-} \\ 
\hline
\multicolumn{2}{|l|}{SSEGCN-BERT~\cite{zhang2022ssegcn}} & \textbf{-\!-\!-} & 81.09 & \textbf{-\!-\!-}  & 77.96 & \textbf{-\!-\!-} & \textbf{-\!-\!-} & \textbf{-\!-\!-} & \textbf{-\!-\!-} \\
\hline
\multicolumn{2}{|l|}{TGCN-BERT~\cite{li2021dual}} & \textbf{-\!-\!-} & 79.95  & \textbf{-\!-\!-}  & 77.03 & \textbf{-\!-\!-} & 82.77 & \textbf{-\!-\!-} & 72.81 \\ 
\hline
\multicolumn{2}{|l|}{Sentic GCN-BERT~\cite{liang2022aspect}} & \textbf{-\!-\!-} & 86.92  & \textbf{-\!-\!-}  & 82.12 & \textbf{-\!-\!-} & 85.32 & \textbf{-\!-\!-} & 91.97 \\ 
\hline
\multicolumn{2}{|l|}{Span-ASTE~\cite{Xu2021SpanLevel}}  & 79.36   & \textbf{-\!-\!-}  & 67.02 & \textbf{-\!-\!-} & 70.47 & \textbf{-\!-\!-} & 74.65 & \textbf{-\!-\!-} \\ 
\hline
\hline
\multicolumn{2}{|l|}{GCAE~\cite{xue2018aspect}} & \textbf{-\!-\!-} & 79.27 & \textbf{-\!-\!-} & 73.56 & \textbf{-\!-\!-} & \textbf{-\!-\!-} & \textbf{-\!-\!-} & \textbf{-\!-\!-}\\ 
\hline
\multicolumn{2}{|l|}{TransCap~\cite{chen2019transfer}} & \textbf{-\!-\!-} & 79.29 & \textbf{-\!-\!-} & 73.87 & \textbf{-\!-\!-} & \textbf{-\!-\!-} & \textbf{-\!-\!-} & \textbf{-\!-\!-}\\ 
\hline
\hline
\multicolumn{2}{|l|}{RoBERTa~\cite{liu2019roberta}} & \textbf{-\!-\!-} & 82.10 & \textbf{-\!-\!-} & 79.73 & \textbf{-\!-\!-} & 62.41 & \textbf{-\!-\!-} & 80.88\\ 
\hline
\multicolumn{2}{|l|}{ASGCN-RoBERTa~\cite{dai2021does}} & \textbf{-\!-\!-} & 80.59 & \textbf{-\!-\!-} & 80.32 &  \textbf{-\!-\!-}& \textbf{-\!-\!-} &  \textbf{-\!-\!-} &  \textbf{-\!-\!-} \\
\hline
\multicolumn{2}{|l|}{RGAT-RoBERTa~\cite{dai2021does}} & \textbf{-\!-\!-} & 81.29 & \textbf{-\!-\!-} & 79.95 &  \textbf{-\!-\!-} &   \textbf{-\!-\!-} &  \textbf{-\!-\!-} &  \textbf{-\!-\!-}\\ 
\hline
\multicolumn{2}{|l|}{PWCN-RoBERTa~\cite{dai2021does}} & \textbf{-\!-\!-} & 80.85  & \textbf{-\!-\!-} & 81.08 &   \textbf{-\!-\!-} & \textbf{-\!-\!-}  & \textbf{-\!-\!-}  & \textbf{-\!-\!-} \\ 
\hline
\multicolumn{2}{|l|}{SARL-RoBERTa~\cite{wang2021eliminating}} & \textbf{-\!-\!-} & 82.44 & \textbf{-\!-\!-} & 82.97 & \textbf{-\!-\!-} & 73.83 & \textbf{-\!-\!-} & 81.92 \\ 
\hline
\multicolumn{2}{|l|}{RoBERTa+MLP~\cite{dai2021does}} & \textbf{-\!-\!-} & 80.96 & \textbf{-\!-\!-} & 80.73 & \textbf{-\!-\!-} & \textbf{-\!-\!-} & \textbf{-\!-\!-} & \textbf{-\!-\!-}  \\ 
\hline
\hline
\multicolumn{2}{|l|}{MN ~\cite{wang2018target}} &\textbf{-\!-\!-}  &64.34  &\textbf{-\!-\!-}  &62.89 & \textbf{-\!-\!-} & \textbf{-\!-\!-} & \textbf{-\!-\!-} & \textbf{-\!-\!-} \\ 
\hline
\multicolumn{2}{|l|}{MN(+AS)~\cite{tang2019progressive}}  &\textbf{-\!-\!-}  &69.15  & \textbf{-\!-\!-} &65.24 & \textbf{-\!-\!-} & \textbf{-\!-\!-} & \textbf{-\!-\!-} & \textbf{-\!-\!-}  \\ 
\hline
\hline
\multicolumn{2}{|l|}{TNet ~\cite{li2018transformation}}  &\textbf{-\!-\!-}  & 71.27 & \textbf{-\!-\!-} &71.75 & \textbf{-\!-\!-} & \textbf{-\!-\!-} & \textbf{-\!-\!-} & \textbf{-\!-\!-} \\ 
\hline
\multicolumn{2}{|l|}{TNet-LF~\cite{li2018transformation}} &  \textbf{-\!-\!-} &71.03  & \textbf{-\!-\!-} &70.14 & \textbf{-\!-\!-} & 59.47 & \textbf{-\!-\!-}& 70.43 \\ 
\hline
\multicolumn{2}{|l|}{TNet-ATT~\cite{tang2019progressive}} &  \textbf{-\!-\!-} &69.44  & \textbf{-\!-\!-} &71.51 & \textbf{-\!-\!-} & \textbf{-\!-\!-} & \textbf{-\!-\!-} & \textbf{-\!-\!-} \\ 
\hline
\multicolumn{2}{|l|}{TNet-AS ~\cite{li2018transformation}} &\textbf{-\!-\!-}  & 71.27 &\textbf{-\!-\!-}  &71.75 & \textbf{-\!-\!-} & \textbf{-\!-\!-} & \textbf{-\!-\!-} & \textbf{-\!-\!-}  \\ 
\hline
\multicolumn{2}{|l|}{TNet-ATT(+AS) ~\cite{tang2019progressive}} &\textbf{-\!-\!-}  & 72.90 &\textbf{-\!-\!-}  &73.84 & \textbf{-\!-\!-} & \textbf{-\!-\!-} & \textbf{-\!-\!-} & \textbf{-\!-\!-}  \\ 
\hline
\hline
\multicolumn{2}{|l|}{LSTM-FC-CNN-LF ~\cite{li2018transformation}} &\textbf{-\!-\!-}  & 70.23 &\textbf{-\!-\!-}  &70.60 & \textbf{-\!-\!-} & \textbf{-\!-\!-} & \textbf{-\!-\!-} & \textbf{-\!-\!-}\\ 
\hline
\multicolumn{2}{|l|}{LSTM+SynATT+TarRep~\cite{nguyen2018effective}} & \textbf{-\!-\!-} & 71.32 & \textbf{-\!-\!-} & 69.23 & \textbf{-\!-\!-} & 66.05 & \textbf{-\!-\!-} & \textbf{-\!-\!-}  \\ 
\hline
\multicolumn{2}{|l|}{LSTM-FC-CNN-AS ~\cite{li2018transformation}} &\textbf{-\!-\!-}  & 70.06 &\textbf{-\!-\!-}  &70.72 & \textbf{-\!-\!-} & \textbf{-\!-\!-} & \textbf{-\!-\!-} & \textbf{-\!-\!-} \\ 
\hline
\multicolumn{2}{|l|}{Sentic-LSTM~\cite{zhou2020sk}} & \textbf{-\!-\!-} & 79.43 & \textbf{-\!-\!-} & 70.88 & \textbf{-\!-\!-} & 79.55 & \textbf{-\!-\!-} & 83.01  \\ 

\hline
\hline
\multicolumn{2}{|l|}{AEN-BERT~\cite{song2019attentional}} & \textbf{-\!-\!-} & 73.76 & \textbf{-\!-\!-} & 76.31 & \textbf{-\!-\!-} & \textbf{-\!-\!-} & \textbf{-\!-\!-} & \textbf{-\!-\!-}   \\ 
\hline
\multicolumn{2}{|l|}{AEN-GloVe~\cite{song2019attentional}} & \textbf{-\!-\!-} & 72.14 & \textbf{-\!-\!-} & 69.04 & \textbf{-\!-\!-} & \textbf{-\!-\!-} & \textbf{-\!-\!-} & \textbf{-\!-\!-}   \\ 
\hline
\end{tabular}
\end{adjustbox}

\end{table*}

\begin{table}[ht]
\centering
\small
\caption{F1 scores of different models which perform the joint task}
\resizebox{0.45\textwidth}{!}{%
\begin{tabular}{|m{3.5cm}|>{\centering\arraybackslash}m{2cm}|>{\centering\arraybackslash}m{2cm}|}
\hline
\textbf{Model} & \textbf{Lap-14} & \textbf{Res-14} \\ \hline
\textbf{InstructABSA~\cite{scaria2023instructabsa}} & 79.34 & 79.47 \\ \hline
\textbf{GRACE~\cite{luo2020grace}} & 70.71 & 77.26 \\ \hline
\textbf{SPAN~\cite{hu2019open}} & 68.06 & 74.92 \\ \hline
\textbf{RACL-BERT~\cite{chen2020relation}} & 63.40 & 75.42\\ \hline
\textbf{BERT-E2E-ABSA~\cite{li2019exploiting}} & 61.12 & 74.72 \\ \hline
\textbf{DOER~\cite{luo2019doer}} & 60.35 & 72.78\\ \hline
\textbf{IMN~\cite{he2019interactive}} & 58.37 & 69.54\\ \hline
\textbf{E2E-TBSA~\cite{li2019unified}} & 57.90 & 69.80 \\ \hline
\end{tabular}%
}
\label{table:f1_scores_new}
\end{table}

\section{Methodology}

In this section, we look on to different approaches we tested out in order to find the most accurate and robust solution. These approaches can be listed below,

\begin{enumerate}
    \item   Fine-Tuning \texttt{LLaMA 2-7B} with Quantized Low Rank Adaptation (QLoRA)
    \item   Fine-Tuning \texttt{Mistral-7B} with Quantized Low Rank Adaptation (QLoRA)
    \item   \texttt{ASGCN+UIKA+Glove} for Sentiment Polarity
    \item   \texttt{SSGCN+Glove} for Sentiment Polarity
    \item   \texttt{Span-ASTE+BERT} for Aspect Extraction
    \item \texttt{SETFIT} for efficient few-shot fine-tuning of Sentence Transformers
    \item \texttt{Instruct-DeBERTa} (Proposed Model)
\end{enumerate}

\subsection{LLaMA 2-7B with QLoRA}
Given the current state-of-the-art interest in Large Language Models (LLMs), we opted to include an LLM-based analysis in our comparative study. 
\texttt{LLaMA 2} is a collection of second-generation open-source LLMs from Meta that comes with a commercial license. \citet{roumeliotis2024llms} presented that \texttt{LLaMA 2} shows a significant leap forward in natural language understanding and generation, by its advanced architecture, large training data, and refined training strategies. The architecture of \texttt{LLaMA 2} is based on the transformer model, a neural network architecture that has proven highly effective in a wide range of NLP tasks. \texttt{LLaMA 2} employs a multi-layered transformer architecture with self-attention mechanisms. It is designed to handle a wide range of natural language processing tasks, with models ranging in scale from 7 billion to 70 billion parameters. 

Fine-tuning in machine learning is the process of adjusting the weights and parameters of a pre-trained model on new data to improve its performance on a specific task. 
There are three main fine-tuning methods in the context:
\begin{enumerate}
    \item  \textbf{Instruction Fine-Tuning (IFT):} According to \citet{peng2023instruction}, IFT involves training the model using prompt completion pairs, showing desired responses to queries.
    \item \textbf{Full Fine Tuning:} Full fine-tuning involves updating all of the weights in a pre-trained model during training on a new dataset, allowing the model to adapt to a specific task. 
    \item \textbf{Parameter-Efficient Fine-Tuning (PEFT):} Selectively updates a small set of parameters, making memory requirements more manageable. There are various ways of achieving Parameter efficient fine-tuning. Low-Rank Parameter (LoRA)~\cite{hu2021lora} and Quantized Low-Ranking Adaptation (QLoRA)~\cite{dettmers2023qlora} are the most widely used and effective.
\end{enumerate}

Traditional fine-tuning of pre-trained language models (PLMs) requires updating all of the model's parameters, which is computationally expensive and requires massive amounts of data; thus making it challenging to attempt on consumer hardware due to inadequate VRAMs and computing. However, Parameter-Efficient Fine-Tuning (PEFT) works by only updating a small subset of the model's most influential parameters, making it much more efficient. Four-bit quantization via QLoRA allows such efficient fine-tuning of huge LLM models on consumer hardware while retaining high performance. QLoRA quantizes a pre-trained language model to four bits and freezes the parameters. A small number of trainable \textit{Low-Rank Adapter} layers are then added to the model. In our case, we created a 4-bit quantization with NF4-type configuration using \texttt{BitsAndBytes}\footURL{https://github.com/TimDettmers/bitsandbytes}. 

According to \citet{dettmers2023qlora} under the model fine-tuning process, Supervised fine-tuning (SFT) is a key step in Reinforcement Learning from Human Feedback (RLHF). The SFT models come with tools to train language models using reinforcement learning, starting with supervised fine-tuning, then reward modelling, and finally, Proximal Policy Optimization (PPO). During this process, we provided the SFT trainer with the model, dataset, LoRA configuration, tokenizer, and training parameters. The model was fine-tuned with a training and evaluation batch size of 4 and for 2 epochs, optimizing its ability to extract aspects and determine sentiment polarity.

To test the fine-tuned model, we used the \textit{Transformers} text generation pipeline including the prompt. The \texttt{LLaMA 2} model was fine-tuned using techniques such as QLoRA, PEFT, and SFT to overcome memory and computational limitations.

\subsection{Mistral-7B with QLoRA}
~\citet{jiang2023mistral} introduced \texttt{Mistral 7B}, a 7-billion-parameter language model designed for superior performance and efficiency. \texttt{Mistral 7B} surpasses the best open 13B model (\texttt{Llama 2}) across all evaluated benchmarks and the best released 34B model (\texttt{Llama 1}) in reasoning, mathematics, and code generation.

\texttt{Mistral 7B} leverages grouped-query attention (GQA) and sliding window attention (SWA). GQA significantly accelerates inference speed and reduces memory requirements during decoding, allowing for higher batch sizes and, consequently, higher throughput—crucial for real-time applications. Additionally, SWA is designed to handle longer sequences more effectively at a reduced computational cost, addressing a common limitation in LLMs. These attention mechanisms collectively enhance the performance and efficiency of \texttt{Mistral 7B}.

The \texttt{Mistral 7B} model was fine-tuned to perform ABSA on the Lap-14 and Res-14 datasets. We have fine-tuned the base model separately for these two datasets and evaluated them separately. This involved customizing the model to better understand and analyze sentiment related to specific aspects within the review texts. The \texttt{Mistral 7B} model was selected from the Hugging Face hub. The mentioned datasets were processed using Pandas to ensure it was in a prompt-compatible format.

To enable efficient training, 4-bit precision loading was configured using BitsAndBytesConfig. We set float16 as the data type for the 4-bit base model, nf4 as quantization type, and nested quantization was disabled to simplify the training process. The tokenizer was loaded and configured to handle padding appropriately. The base model was then loaded with the quantization configuration, ensuring it was prepared for low-bit precision training.

In our fine-tuned models we set Attention dimension to  64, Scaling parameter to 64, and Dropout probability: 0.1 for efficient Low-Rank Adaptation (LoRA) parameters. Then the fine-tuning was conducted using the SFTTrainer with the defined training arguments. The dataset was loaded, and the model underwent supervised fine-tuning, adjusting to the specific requirements of ABSA on the mentioned datasets. As the final stage the post-training, the model was saved and reloaded in FP16 precision. The LoRA weights were merged back into the base model to create a final, streamlined version suitable for deployment. We fine-tuned this model using a batch size of 4 for both training and evaluation over 2 epochs, enhancing its capability to extract aspects and identify sentiment polarity. 

To test the fine-tuned model,  we developed a function to process user input and generate corresponding aspects and sentiments. The function takes an input sentence and utilizes a text generation pipeline where we set the prompt and specific parameters for sampling.

By utilizing Hugging Face libraries such as \texttt {transformers}\footURL{https://huggingface.co/transformers/}, \texttt{accelerate}\footURL{https://huggingface.co/accelerate/}, \texttt{peft}\footURL{https://huggingface.co/peft/}, \texttt{trl}\footURL{https://huggingface.co/trl/}, and \texttt{bitsandbytes}, we were able to successfully fine-tune and evaluate the both 7B parameter \texttt{LLaMA 2} model and \texttt{Mistral} model on a consumer GPU.

\subsection{SetFit}
Few-shot learning has become increasingly essential in addressing label-scarce scenarios, where data annotation is often time-consuming and expensive. These methods aim to adapt pre-trained language models (PLMs) to specific downstream tasks using only a limited number of labelled training examples.
One of the primary obstacles is the reliance on large-scale language models, which typically contain billions of parameters, demanding substantial computational resources and specialized infrastructure. Moreover, these methods frequently require manual crafting of prompts, introducing variability and complexity in the training process, thus restricting accessibility for researchers and practitioners.


In response to this, \citet{tunstall2022efficient} proposed \texttt{SETFIT} (Sentence Transformer Fine-tuning) which presents an innovative framework for efficient and prompt-free few-shot fine-tuning of Sentence Transformers (ST). Diverging from existing methods, \texttt{SETFIT} does not necessitate manually crafted prompts and achieves high accuracy with significantly fewer parameters. Through a straightforward yet effective approach, \texttt{SETFIT} simplifies the few-shot learning process, making it more accessible and practical across various applications. \texttt{SETFIT} is a novel framework for efficient few-shot fine-tuning of ST without the need for manual prompts. The \texttt{SETFIT} approach consists of two main steps: 

\begin{enumerate}
    \item  Fine-tuning a pretrained ST on a small number of text pairs in a contrastive Siamese manner.
    \item Training a classification head using the resulting ST to generate rich text embeddings.
\end{enumerate}

In the first step, the ST is fine-tuned using a contrastive loss function, which encourages the model to learn discriminative representations of similar and dissimilar text pairs. 
In the second step, a simple classification head is trained on top of the fine-tuned ST to perform downstream tasks such as text classification or similarity ranking. By decoupling the fine-tuning and classification steps, \texttt{SETFIT} achieves high accuracy with orders of magnitude fewer parameters than existing methods, making it computationally efficient and scalable.

\subsection{Instruct-DeBERTa (Proposed Model)}

In this study, we developed an aspect-based sentiment analysis pipeline utilizing transformer-based models to automatically extract aspects and analyze sentiments in textual data. The pipeline is composed of two primary stages: aspect extraction and sentiment classification. For these two stages, we utilized the best models for each task that we found through our thorough literature review which is being summarized in \autoref{tab:tablenewlitreview}. When looking on to the analysis it is clear that \texttt{InstructABSA}~\cite{scaria2023instructabsa} performs the best in all the analyzed datasets irrespective of the domain. For the Res-14 dataset, it recorded an F1 score of 92.10\% which was higher than all the other reported models. It still remained the highest on the Res-15 data set and was only 1.67\% less than the highest recorded accuracy under the Res-16 dataset. But \texttt{P-SUM}~\cite{karimi2020improving} which reported the highest F1 score for the Res-16 dataset performed significantly less than \texttt{InstructABSA} in the previous datasets. Hence \texttt{InstructABSA} still remained the best option for aspect extraction. Moreover, for the Lap-14 dataset, \texttt{InstructABSA} topped the charts again showcasing the models' adaptability and robustness regardless of the relevant domain. Hence \texttt{InstructABSA} was selected as the best performing model for aspect extraction. When looking at the performances on the sentiment polarity task \texttt{DeBERTa-V3-baseabsa-V1}~\cite{DBLP:conf/cikm/0008ZL23,YangZMT21} is the best overall performing model across all the datasets. It showcases an accuracy of 90.94\% for the Res-14 dataset which is the highest overall. It shows the same promising results in the Res-15 and Res-16 datasets. In addition to that  \texttt{DeBERTa-V3-baseabsa-V1} also shows the adaptability of the model by recording the highest accuracy for the Lap-14 dataset as well which falls in to a complete different domain.

Since we identified the best overall performing models for individual tasks of aspect extraction and sentiment polarity detection, we tried to exploit the performances of these models and build up a hybrid model that performs the joint task of aspect extraction and sentiment polarity detection by itself.
Then we created a \textbf{novel hybrid model} utilizing these models to build up a model which as per our research gives the best performance as a pipelined hybrid model, which in fact makes this the SOTA model for the pipelined aspect extraction and sentiment polarity classification  task also known as the joint task for ABSA. 

\begin{figure*}[htbp]
    \centering
    \includegraphics[width=\textwidth]{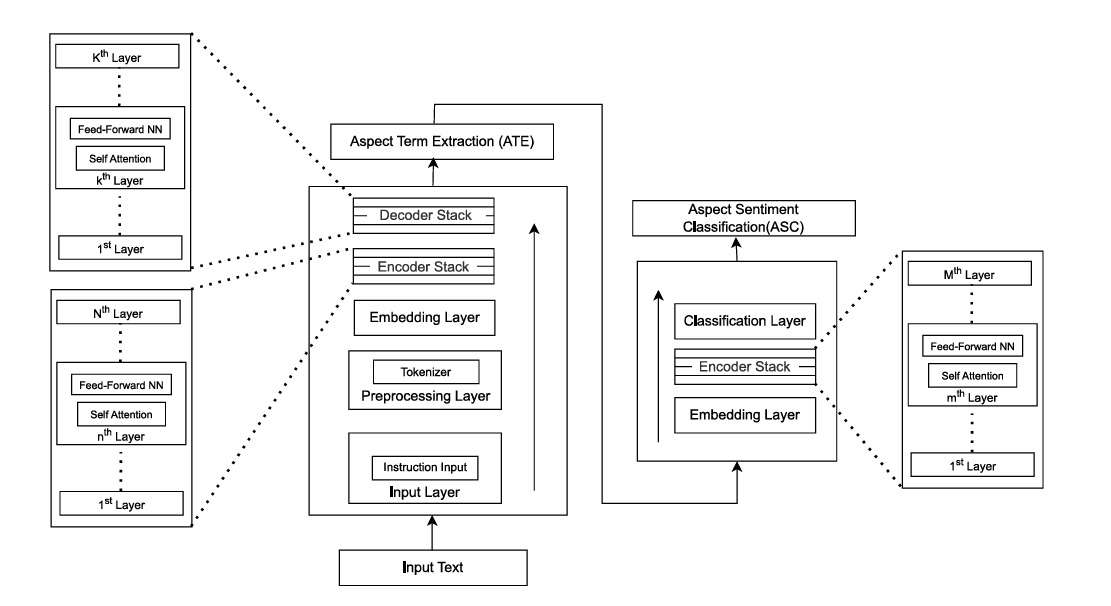}
    \caption{Proposed model}
    \label{fig:our_model}
\end{figure*}

\autoref{fig:our_model} shows the proposed model of our study. The model structure used for 
\textbf{ATE} is \texttt{\textbf{InstructABSA}} while the model used for \textbf{ASC} is \texttt{\textbf{DeBERTa-V3-baseabsa-V1}}. The collective model is named as \texttt{\textbf{Instruct-DeBERTa}}, which stands for \texttt{\textbf{Instruct}ABSA} for aspect term extraction and \texttt{\textbf{DeBERTa}-V3-baseabsa-V1} for aspect sentiment classification. When looking on to \autoref{fig:our_model} it shows how these two independent models are being pipelined to build up a single joint task model.


The algorithm for our proposed model can be presented as below for further clarification.

\hspace{1 cm}

\begin{algorithm}
\caption{Aspect-based Sentiment Analysis (ABSA)}
\label{algo:absa}
\begin{algorithmic}[1]
\REQUIRE Review $X$
\STATE \textbf{Aspect Term Extraction (ATE):}
\STATE \quad $A_c = f_{ATE}(X)$
\STATE \textbf{Target Aspect Filtering:}
\STATE \quad $A = f_{filter}(A_c)$
\STATE \textbf{Aspect Sentiment Classification (ASC):}
\STATE \quad $S = \{\}$
\FOR{each aspect term $a$ in $A$}
    \STATE \quad $s = f_{ASC}(X, a)$
    \STATE \quad Add $(a, s)$ to $S$: $S = S \cup \{(a, s)\}$
\ENDFOR
\RETURN Final aspect terms $A$ with sentiment labels $S$: $\{(a, s) \mid a \in A, s = f_{ASC}(X, a)\}$
\end{algorithmic}
\end{algorithm}

Sets:
\begin{itemize}
  \item $X$: Review represented as a word sequence ($X = \{x_1, x_2, \ldots, x_n\}$)
  \item $A$: Set of extracted aspect terms ($A \subseteq X$)
  \item $S$: Set of sentiment labels for aspect terms ($S \subseteq \{positive, negative, neutral\}$)
\end{itemize}

Functions:
\begin{itemize}
  \item $f_{ATE}(X)$: Function for Aspect Term Extraction. Takes review $X$ and returns candidate terms ($A_c$). ($A_c \subseteq X$)
  \item $f_{filter}(A_c)$:A filtering function. Takes candidate terms and returns refined aspects ($A$). ($A \subseteq A_c$)
  \item $f_{ASC}(X,a)$: Function for Aspect Sentiment Classification. Takes review $X$ and an aspect term $a$, returns sentiment label $s$. ($s \in \{positive, negative, neutral\}$)
\end{itemize}

\section{Results}

\autoref{tab:Results_F1} presents the F1 scores for the models being built and evaluated by ourselves. In addition to that we have presented \autoref{fig:aspect} and \autoref{fig:sentiment_polarity} which give a view on the robustness of each model for the aspect term extraction task and the sentiment polarity task separately based on the accuracies. 
According to \autoref{fig:aspect} and \autoref{fig:sentiment_polarity}, if the model shows high accuracy for each task and the accuracies for the two domains do not exhibit drastic deviations, then the relevant model will be selected.

\subsection{LLaMA 2-7B with QLoRA}
The first section of \autoref{tab:Results_F1} shows the performance of \texttt{Llama-2-7b}~\cite{touvron2023llama} with QLoRA~\cite{dettmers2023qlora}. It emphasizes that this model shows notable performance in both aspect extraction and sentiment polarity tasks. On Res-14 and Lap-14 datasets, \texttt{Llama 2} shows a slight edge in aspect extraction compared to sentiment polarity. 
These performances were obtained using the L4 GPU emphasizing the model's efficiency and effectiveness in computational performance. 

\subsection{Mistral 7B with QLoRA}
According to the values \autoref{tab:Results_F1}, the model \texttt{Mistral 7B} exhibits superior performance in both aspect extraction and sentiment polarity tasks compared to the \texttt{Llama 2} model. Specifically, \texttt{Mistral 7B} achieves higher F1 scores across both Res-14 and Lap-14 datasets, indicating its greater capability in accurately identifying aspects and determining sentiment polarity within the text. These results were achieved using an L4 GPU, similar to the \texttt{Llama 2} model.

When comparing the two models, \texttt{Mistral 7B} demonstrates a clear advantage in both tasks. While \texttt{Llama 2} performs competently, \texttt{Mistral 7B} consistently outperforms it, showcasing its enhanced effectiveness and reliability in handling aspect extraction and sentiment polarity analysis. This comparison highlights \texttt{Mistral 7B}'s performance, making it a more capable model for these specific natural language processing tasks.

\subsection{Some models with BERT and GloVe}
As a part of the comparative study for the survey, we conducted experiments using several advanced models for aspect-based sentiment analysis. We experimented three main models: \texttt{SSGCN+Glove}, \texttt{ASGCN+UIKA+Glove}\cite{zhang-etal-2019-aspect}, and \texttt{Span-ASTE+BERT} \cite{Xu2021SpanLevel} both in local and Colab environments. The focus was on to evaluating their performance on the SemEval 2014 dataset. These results are stated in \autoref{tab:Results_F1}. But still we observed that \texttt{Instruct-DeBERTa} outperforms all.

\subsection{SetFit}

In the second section of \autoref{tab:Results_F1}, we provide a comprehensive overview of the F1-scores for attained by various sentence models using the \texttt{SETFIT} framework~\cite{tunstall2022efficient} for the same amount of selected data in the respective stated data sets. 
If a model is reported in a single row, it means we have used the said sentence transformer model for both aspect extraction and sentiment polarity identification (eg, \texttt{BGE}~\cite{xiao2023c}). In the cell blocks where a model is followed by other models with $+$ are combinations. For example, the first row of \texttt{Paraphrase-MiniLM-L6-v2}~\cite{wang2020minilm,reimers2019sentence} contains results of that model being used both for aspect extraction and sentiment polarity identification. The subsequent line with \texttt{+MpNet}~\cite{song2020mpnet} indicates that \texttt{Paraphrase-MiniLM-L6-v2} was used for the aspect extraction component and \texttt{MpNet} was used for the sentiment polarity identification component. 

At this point, a question may be raised as to why would the aspect extraction have two different values for accuracy in \autoref{fig:aspect} and \autoref{fig:sentiment_polarity} in the two configurations if in both cases the same model (ie, \texttt{Paraphrase-MiniLM-L6-v2} in this example) was used for that task. The reason is the fact that the fine-tuning is conducted end-to-end in a holistic manner and thus, the choice of the model used for the sentiment polarity identification ends up influencing the ultimate accuracy obtained by the aspect extraction component. It may enhance the result as in the case of \texttt{Paraphrase-MiniLM-L6-v2} and \texttt{MpNet}. It may also hinder as in the case of \texttt{LaBSE}. 

Overall, it can be noted that \texttt{LaBSE}~\cite{feng2022language} consistently emerges as a standout performer; either by itself or as the aspect extraction component of a pair. It can be argued that this robust performance is owed to its capability to capture nuanced complex information crucial for understanding both aspect-based sentiment analysis and sentiment polarity classification tasks.
Specifically on the sentiment polarity classification task, it can be noted that \texttt{Mpnet} and \texttt{RoBERTa-STSb-v2}~\cite{cer2017semeval} elevates performance multiple configurations. 

To give a better overview of how various models perform, we include \autoref{fig:aspect} and \autoref{fig:sentiment_polarity}, which visualize the \texttt{SetFit} accuracy values for the models that we evaluated. 
In \autoref{fig:aspect}, we present a detailed analysis of aspect extraction accuracy for various models. \texttt{LaBSE} emerging as the top performer across both datasets can easily be noted. It is also evident how \texttt{ALBERT+DistilRoBERTa} and \texttt{LaBSE+RoBERTa-STSb} closely follow with accuracies verging on 90\%. 
Similarly, in \autoref{fig:sentiment_polarity}, we look into the analysis of sentiment polarity identification percentages for the same models and datasets. Here, \texttt{LaBSE+RoBERTa-STSb} shows the highest accuracy for Res-14 while \texttt{LaBSE+MpNet} shows the highest accuracy for Lap-14. 

\subsection{Instruct-DeBERTa (Proposed Model)}

\begin{figure}[h!]
    \centering
    \begin{adjustbox}{width=\linewidth}
    \begin{tikzpicture}
    \begin{axis}[
        xlabel={},
        ylabel={Sentiment Polarity Identification Percentage (\%)},
        xtick=data,
        xticklabels={
            Llama-2-7b~\cite{touvron2023llama} with QLoRA~\cite{dettmers2023qlora},
            Mistral-7b~\cite{jiang2023mistral} with QLoRA~\cite{dettmers2023qlora},
            Instruct-DeBERTa (Our Model),
            BGE~\cite{xiao2023c} (Small),
            Sentence-T5~\cite{ni2022sentence} (Base),
            RoBERTa-STSb-v2~\cite{cer2017semeval,reimers2019sentence} (Base),
            MpNet~\cite{song2020mpnet},
            Paraphrase-MiniLM-L6-v2~\cite{wang2020minilm,reimers2019sentence},
            Paraphrase-MiniLM-L6-v2~\cite{wang2020minilm,reimers2019sentence}+MpNet~\cite{song2020mpnet},
            CLIP-ViT-B-32-multilingual-v1~\cite{radford2021learning,reimers2019sentence},
            facebook-dpr-ctx-encoder-multiset-base~\cite{karpukhin-etal-2020-dense},
            SPECTER~\cite{specter2020cohan},
            GTR~\cite{ni2022large} (Base),
            SBERT~\cite{reimers2019sentence} (Base),
            TinyBERT~\cite{jiao2020tinybert,reimers2019sentence},
            ALBERT~\cite{lan2019albert,reimers2019sentence},
            ALBERT~\cite{lan2019albert,reimers2019sentence}+DistilRoBERTa~\cite{Sanh2019DistilBERTAD},
            DistilRoBERTa~\cite{Sanh2019DistilBERTAD},
            DistilRoBERTa~\cite{Sanh2019DistilBERTAD}+All-MiniLM-L6-v2~\cite{wang2020minilm,reimers2019sentence},
            LaBSE~\cite{feng2022language},
            LaBSE~\cite{feng2022language}+MpNet~\cite{song2020mpnet},
            LaBSE~\cite{feng2022language}+GTR~\cite{ni2022large} (Base),
            LaBSE~\cite{feng2022language}+RoBERTa-STSb-v2~\cite{cer2017semeval,reimers2019sentence} (Base)
        },
        x tick label style={rotate=90,anchor=east,font=\footnotesize},
        y tick label style={font=\footnotesize},
        ytick={45,50, 55, 60, 65, 70, 75, 80, 85, 90,95,100},
        ymin=45,
        ymax=100,
        width=1.2\textwidth, 
        height=10cm, 
        grid=both,
        axis lines=left,
        axis line style={-}, 
        legend style={at={(0.5,-0.7)},anchor=north,font=\footnotesize,draw=none},
        legend columns=3,
        enlarge x limits=0.02,
        ylabel style={font=\scriptsize},
        xlabel style={font=\footnotesize}
    ]
    
    \addplot[mark=*,blue,only marks] coordinates {
        (1, 68.01)
        (2, 75.33)
        (3, 88.96)
        (4, 73.20)
        (5, 77.90)
        (6, 78.70)
        (7, 77.95)
        (8, 62.00)
        (9, 79.50)
        (10, 69.30)
        (11, 79.52)
        (12, 71.60)
        (13, 72.40)
        (14, 70.86)
        (15, 72.40)
        (16, 71.65)
        (17, 75.50)
        (18, 73.20)
        (19, 71.60)
        (20, 76.40)
        (21, 74.80)
        (22, 74.00)
        (23, 80.30)
    };
    
    \addplot[mark=*,red,only marks] coordinates {
        (1, 68.62)
        (2, 68.62)
        (3, 90.12)
        (4, 74.80)
        (5, 71.60)
        (6, 66.10)
        (7, 70.07)
        (8, 61.40)
        (9, 70.00)
        (10, 52.60)
        (11, 68.50)
        (12, 49.60)
        (13, 74.00)
        (14, 59.05)
        (15, 62.90)
        (16, 65.35)
        (17, 66.90)
        (18, 66.10)
        (19, 65.30)
        (20, 65.40)
        (21, 75.60)
        (22, 73.20)
        (23, 70.10)
    };
    
    \legend{Res-14 ,Lap-14}
    
    \end{axis}
    
    \draw [black] (current axis.south west) rectangle (current axis.north east);
    \end{tikzpicture}
    \end{adjustbox}
    \caption{Sentiment polarity percentage of models}
    \label{fig:sentiment_polarity}
\end{figure}
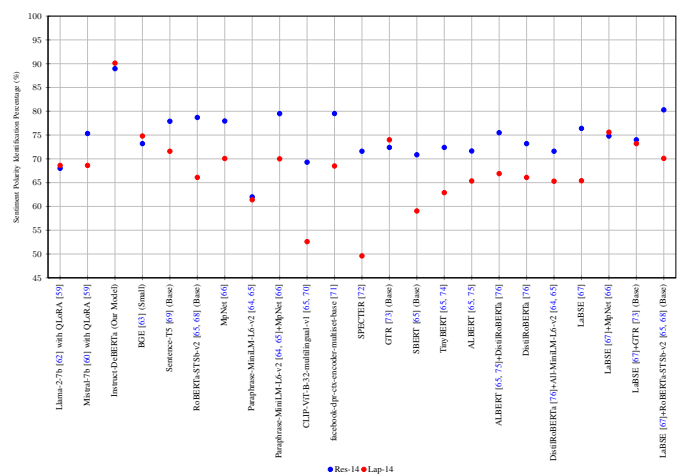

From all the evaluated models, our model performs the best with the highest accuracies and F1 scores. It is also robust for both domains which makes it the best performing model. As discussed in the methodology we selected the best performing models from \autoref{tab:tablenewlitreview} to create our own hybrid model. When looking at \autoref{tab:Results_F1}, \autoref{fig:aspect} and \autoref{fig:sentiment_polarity} it is clear that \texttt{Instruct-DeBERTa} outperforms our finetuned Llama, Mistral, and all the Setfit based models. 


 \autoref{last_table} shows how the two best models we selected for each subtask perform individually on their relevant task. These F1 scores are for the combined task, which means our model is capable of performing both the aspect extraction and the sentiment polarity tasks. For the first task which is extracting the aspects, our model gives closer accuracies for what has been reported by \citet{scaria2023instructabsa} for the \texttt{InstructABSA} model. Different ways of splitting a dataset can affect the reported accuracies. Also for the sentiment polarity classification task the original model, \texttt{DeBERTa-V3-base-absa-v1} by ~\citet{DBLP:conf/cikm/0008ZL23,YangZMT21} which is specialized only for detecting polarities gives slightly higher accuracies than our hybrid model. 
As seen for the Res-14 dataset the sentiment polarity accuracy for the individual task by \texttt{DeBERTa-V3-base-absa-v1} is reported as 90.94\% while our reported 88.63\%. This is due to the models being pipelined and the extracted aspects from the first model is being fed to the second model rather than calculating the accuracies separately for individual tasks. Hence the slight deduction in the hybrid model is justified.

So, looking on to all the past models in \autoref{tab:tablenewlitreview} and the models that we worked on in \autoref{tab:Results_F1}, it is clear that our model, \texttt{Instruct-DeBERTa} is the best performing hybrid model designed for the combined task of aspect extraction and sentiment polarity detection. 
Moreover, our hybrid model shows promising results in the laptop domain as well. Our model gives an F1-score of 91.56\% and 89.65\% for aspect extraction and sentiment polarity respectively.

\begin{figure}
    \centering
    \includegraphics[width=1.0\linewidth]{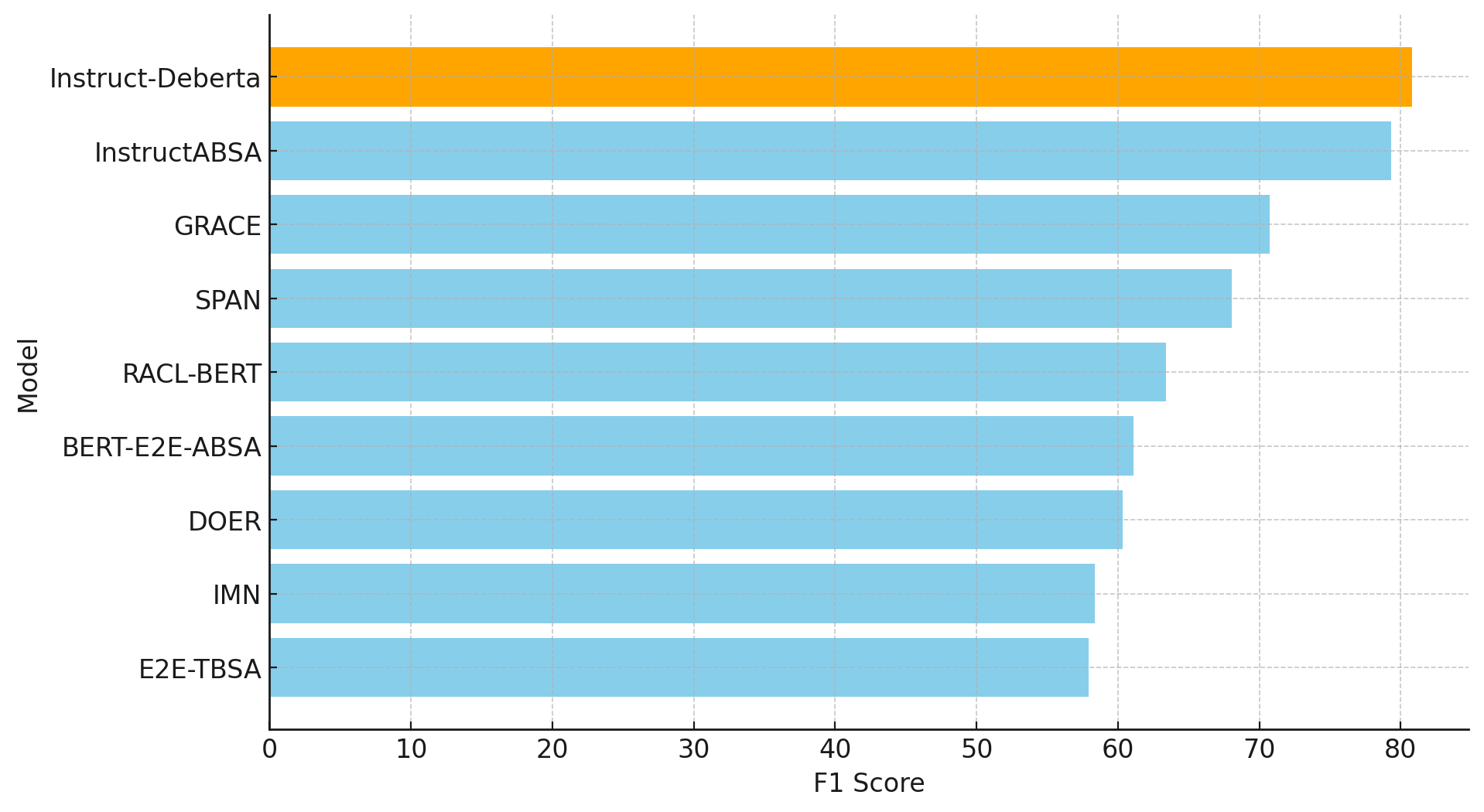}
    \caption{F1 scores of models for the joint task of ASC and ATE for lap-14}
    \label{fig:lap14new}
\end{figure}

\begin{figure}
    \centering
    \includegraphics[width=1.0\linewidth]{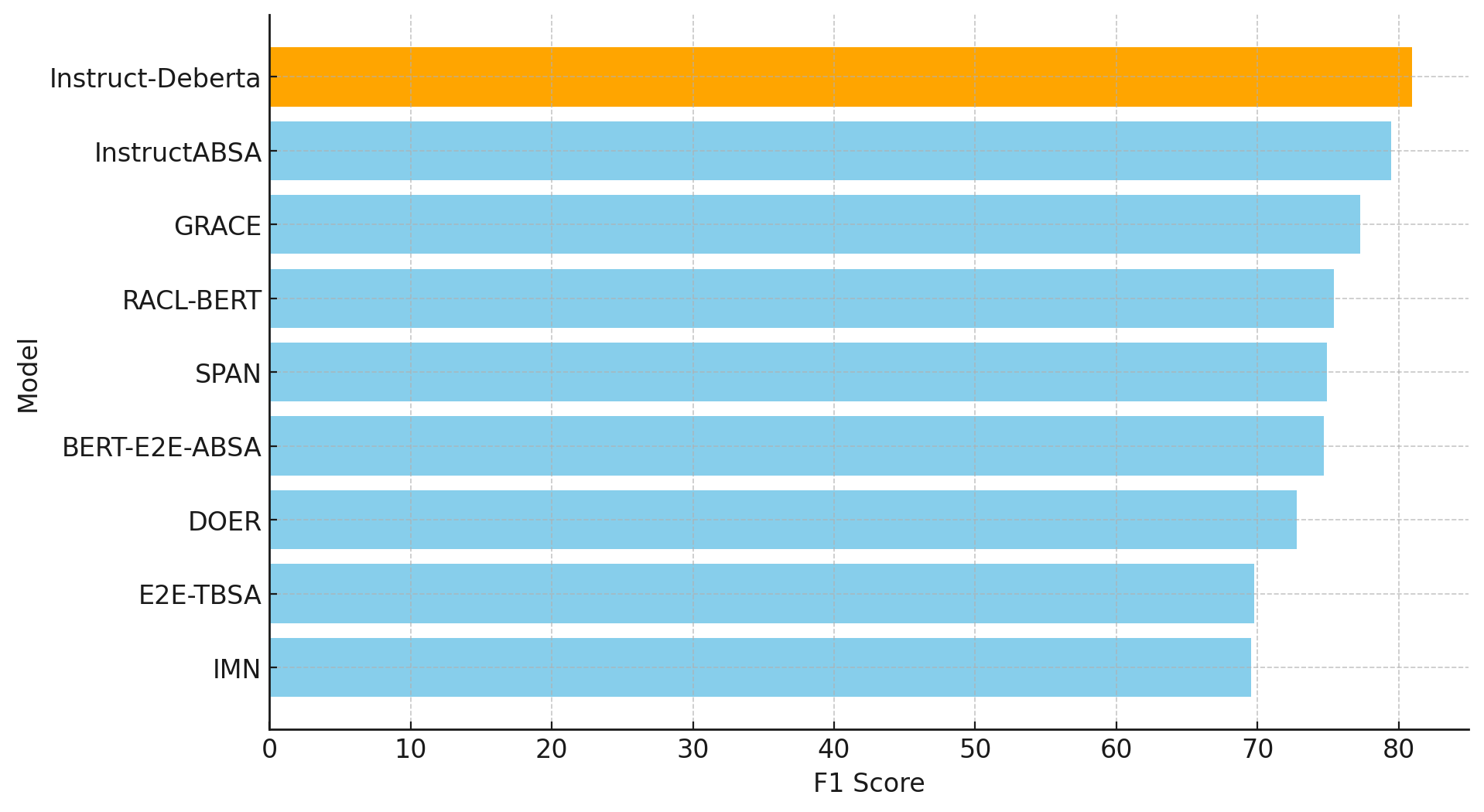}
    \caption{F1 scores of models for the joint task of ASC and ATE for res-14}
    \label{fig:rest14new}
\end{figure}

Also \autoref{table:f1_scores_new}, in the literature review lists out the joint models which are equivalent to the model we built. These perform the joint task of ABSA. Here in order for the F1 score to be counted the aspect and the respective sentiment in the original dataset needs to be correct.
The F1 scores of these models along with our model can be visualized in Figure \ref{fig:lap14new} and Figure \ref{fig:rest14new}. It is clear that our model clearly out performs the currently available joint task hybrid models. It gives a pair extraction F1 score of 80.78\% and 80.94\% which exceed the current reported highest accuracy for the rest-14 and lap-14 datasets.
From \autoref{table:f1_scores_new}, \autoref{fig:lap14new} and \autoref{fig:rest14new} it is clear that our model is the best performing joint task model. Our model outperforms all other hybrid models in both domains which again proves that it is not only accurate but also robust to different domains as well.

\begin{table*}[!htbp]
\caption{F1 scores of models evaluated by this study on the SemEval 2014~\cite{pontiki-etal-2014-semeval} benchmark}
\begin{adjustbox}{width=\textwidth,center,scale={1.0}}
\begin{tabular}{|l|l|c|c|c|c|}
\hline
\multicolumn{2}{|l|}{\multirow{3}{*}{\textbf{Model}}} & \multicolumn{4}{c|}{\textbf{F1 Score (\%)}} \\
\hhline{~~----}
\multicolumn{2}{|l|}{} & \multicolumn{2}{c|}{\textbf{Res-14}}& \multicolumn{2}{c|}{\textbf{Lap-14}} \\
\hhline{~~----}
\multicolumn{2}{|l|}{} & \textbf{Aspect Extraction} & \textbf{Sentiment Polarity}& \textbf{Aspect Extraction} & \textbf{Sentiment Polarity} \\
\hline
\multicolumn{2}{|l|}{Llama-2-7b~\cite{touvron2023llama} with QLoRA~\cite{dettmers2023qlora}} & 71.94 &69.29  & 71.66 & 66.53 \\ 
\hline
\multicolumn{2}{|l|}{Mistral-7b~\cite{jiang2023mistral} with QLoRA~\cite{dettmers2023qlora}} & 81.33 & 76.46 &77.65  & 72.40 \\ 
\hline
\multicolumn{2}{|l|}{Instruct-DeBERTa (Proposed Model)} & \textbf{91.39} & \textbf{88.63} & \textbf{91.56} & \textbf{89.65} \\ 
\hline
\multicolumn{2}{|l|}{ASGCN+GloVe+UIKA~\cite{liu2023unified}} & \textbf{-\!-\!-} & 76.93  & \textbf{-\!-\!-} & 75.21 \\ 
\hline
\multicolumn{2}{|l|}{SSGCN+Glove~\cite{zhang2022ssegcn}} & \textbf{-\!-\!-} & 76.43& \textbf{-\!-\!-} & 76.17 \\ 
\hline
\multicolumn{2}{|l|}{SPAN-ASTE~\cite{Xu2021SpanLevel}}  & 67.54 & \textbf{-\!-\!-} & 61.12 & \textbf{-\!-\!-} \\ 
\hline
\parbox[t]{2mm}{\multirow{17}{*}{\rotatebox[origin=c]{90}{ \texttt{SETFIT}~\cite{tunstall2022efficient}}}} & BGE~\cite{xiao2023c} (Small)  & 72.24 & 75.59 & 64.79 & 75.59 \\
\hhline{~-----}
& Sentence-T5~\cite{ni2022sentence} (Base) & 56.82 & 78.74 & 63.29 & 74.01 \\ 
\hhline{~-----}
&   RoBERTa-STSb-v2~\cite{cer2017semeval,reimers2019sentence} (Base) & 82.37 & 77.95 & 84.26 & 67.71 \\ 
\hhline{~-----}
& Paraphrase-MiniLM-L6-v2~\cite{wang2020minilm,reimers2019sentence} & 84.58 & 71.65 & 83.14 & 64.56 \\ 
& ~~+MpNet~\cite{song2020mpnet} & 84.58 & 78.74 & 79.40 & 70.07 \\
\hhline{~-----}
& CLIP-ViT-B-32-multilingual-v1~\cite{radford2021learning,reimers2019sentence}   & 81.49 & 59.05 & 73.03 & 53.54 \\ 
\hhline{~-----}
& facebook-dpr-ctx-encoder-multiset-base ~\cite{karpukhin-etal-2020-dense}  & 76.54 & 78.74 & 75.52 & 77.45  \\  
\hhline{~-----}
& SPECTER~\cite{specter2020cohan}  & 81.93 & 71.65 & 77.52 & 55.11 \\  
\hhline{~-----}
& GTR~\cite{ni2022large} (Base) & 81.85 &74.80  & 84.70 & 74.80 \\ 
\hhline{~-----}
& 
SBERT~\cite{reimers2019sentence} (Base) & 83.18 & 70.86 & 84.32 & 61.41 \\ 
\hhline{~-----}
&
TinyBERT~\cite{jiao2020tinybert,reimers2019sentence} & 78.76 & 73.22 & 82.83  & 63.77 \\ 
\hhline{~-----}
& ALBERT~\cite{lan2019albert,reimers2019sentence}  & 80.08 & 74.80 & 80.22  & 66.92  \\ 
& ~~+DistilRoBERTa~\cite{Sanh2019DistilBERTAD}  & 81.49 & 74.81 & 79.40 & 68.50 \\ 
\hhline{~-----}
& DistilRoBERTa~\cite{Sanh2019DistilBERTAD} & 84.95 & 75.59  & 80.97  & 69.29 \\ 
& ~~+All-MiniLM-L6-v2~\cite{wang2020minilm,reimers2019sentence} & 85.46  & 71.65 & 81.27  & 63.77 \\  
\hhline{~-----}
& MpNet~\cite{song2020mpnet} & 86.28 & 77.95 & 88.80 & 73.22 \\ 
\hhline{~-----}
& LaBSE~\cite{feng2022language} & 89.38 & 73.23  & 90.30 & 64.57  \\ 
& ~~+MpNet~\cite{song2020mpnet} & 88.55  & 74.80 & 89.51 & 75.59 \\ 
& ~~+GTR~\cite{ni2022large} (Base) & 88.55  &74.02  &87.27  & 73.22 \\ 
& ~~+RoBERTa-STSb-v2~\cite{cer2017semeval,reimers2019sentence} (Base) & 90.30 & 77.17 & 89.51 &70.08 \\ 
\hline
\end{tabular}
\end{adjustbox}
\label{tab:Results_F1}
\end{table*}

\begin{table*}[!htbp]
\caption{F1 scores for the selected models individually and when pipe-lined}.
\begin{adjustbox}{width=\textwidth,center}
\scriptsize
\begin{tabular}{|l|l|c|c|c|c|c|c|c|c|}
\hline
\multicolumn{2}{|l|}{\multirow{3}{*}{\textbf{Model}}} & \multicolumn{8}{c|}{\textbf{F1 Score (\%)}} \\
\hhline{~~--------}
\multicolumn{2}{|l|}{} & \multicolumn{2}{c|}{\textbf{Res-14}} & \multicolumn{2}{c|}{\textbf{Lap-14}} & \multicolumn{2}{c|}{\textbf{Res-15}} & \multicolumn{2}{c|}{\textbf{Res-16}} \\
\hhline{~~--------}
\multicolumn{2}{|l|}{} & \textbf{AE} & \textbf{SP} & \textbf{AE} & \textbf{SP} & \textbf{AE} & \textbf{SP} & \textbf{AE} & \textbf{SP} \\
\hline
\multicolumn{2}{|l|}{\textbf{InstructABSA}~\cite{scaria2023instructabsa}} & \textbf{92.10} & \textbf{-\!-\!-} & \textbf{92.30} & \textbf{-\!-\!-} & \textbf{76.64} & \textbf{-\!-\!-} & \textbf{80.32} & \textbf{-\!-\!-}   \\ 
\hline
\multicolumn{2}{|l|}{\textbf{DeBERTa-V3-base-absa-v1.1$^*$}~\cite{DBLP:conf/cikm/0008ZL23,YangZMT21} } & \textbf{-\!-\!-} & \textbf{90.94} & \textbf{-\!-\!-} & \textbf{90.32} & \textbf{-\!-\!-} & \textbf{89.55} & \textbf{-\!-\!-} & \textbf{83.71}\\ 
\hline
\multicolumn{2}{|l|}{Instruct-DeBERTa (Proposed Model)} & 91.39 & 88.63 & 91.56 & 89.65 & 75.13 & 81.26 & 77.79 & 79.35  \\ 
\hline
\end{tabular}
\end{adjustbox}
\label{last_table}
\end{table*}

\begin{figure}[htbp]
    \centering
    \begin{adjustbox}{width=\linewidth}
    \begin{tikzpicture}
    \begin{axis}[
        xlabel={},
        ylabel={Aspect Extraction Accuracy (\%)},
        xtick=data,
        xticklabels={
            Llama-2-7b~\cite{touvron2023llama} with QLoRA~\cite{dettmers2023qlora},
            Mistral-7b~\cite{jiang2023mistral} with QLoRA~\cite{dettmers2023qlora},
            Instruct-DeBERTa (Our Model),
            BGE~\cite{xiao2023c} (Small),
            Sentence-T5~\cite{ni2022sentence} (Base),
            RoBERTa-STSb-v2~\cite{cer2017semeval,reimers2019sentence} (Base),
            MpNet~\cite{song2020mpnet},
            Paraphrase-MiniLM-L6-v2~\cite{wang2020minilm,reimers2019sentence},
            Paraphrase-MiniLM-L6-v2~\cite{wang2020minilm,reimers2019sentence}+MpNet~\cite{song2020mpnet},
            CLIP-ViT-B-32-multilingual-v1~\cite{radford2021learning,reimers2019sentence},
            facebook-dpr-ctx-encoder-multiset-base ~\cite{karpukhin-etal-2020-dense},
            SPECTER~\cite{specter2020cohan},
            GTR~\cite{ni2022large} (Base),
            SBERT~\cite{reimers2019sentence} (Base),
            TinyBERT~\cite{jiao2020tinybert,reimers2019sentence},
            ALBERT~\cite{lan2019albert,reimers2019sentence},
            ALBERT~\cite{lan2019albert,reimers2019sentence}+DistilRoBERTa~\cite{Sanh2019DistilBERTAD},
            DistilRoBERTa~\cite{Sanh2019DistilBERTAD},
            DistilRoBERTa~\cite{Sanh2019DistilBERTAD}+All-MiniLM-L6-v2~\cite{wang2020minilm,reimers2019sentence},
            LaBSE~\cite{feng2022language},
            LaBSE~\cite{feng2022language}+MpNet~\cite{song2020mpnet},
            LaBSE~\cite{feng2022language}+GTR~\cite{ni2022large} (Base),
            LaBSE~\cite{feng2022language}+RoBERTa-STSb-v2~\cite{cer2017semeval,reimers2019sentence} (Base)
        },
        x tick label style={rotate=90,anchor=east,font=\footnotesize},
        y tick label style={font=\footnotesize},
        ytick={45,50, 55, 60, 65, 70, 75, 80, 85, 90,95,100},
        ymin=45,
        ymax=100,
        width=\textwidth, 
        height=8cm, 
        grid=both,
        axis lines=left,
        axis line style={-}, 
        legend style={at={(0.5,-0.4)},anchor=north,font=\footnotesize,yshift=-90pt,draw = none},
        legend columns=3,
        enlarge x limits=0.02,
        ylabel style={font=\scriptsize},
        xlabel style={font=\footnotesize}
    ]
    
    \addplot[mark=*,blue,only marks] coordinates {
        (1, 71.01)
        (2, 80.08)
        (3, 91.55)
        (4, 60.10)
        (5, 78.70)
        (6, 79.20)
        (7, 87.16)
        (8, 79.80)
        (9, 85.40)
        (10, 81.90)
        (11, 82.30)
        (12, 81.90)
        (13, 82.30)
        (14, 82.74)
        (15, 83.10)
        (16, 76.99)
        (17, 84.50)
        (18, 85.00)
        (19, 85.90)
        (20, 90.30)
        (21, 88.50)
        (22, 88.50)
        (23, 88.50)
    };
    
    \addplot[mark=*,red,only marks] coordinates {
        (1, 75.94)
        (2, 75.94)
        (3, 91.76)
        (4, 86.50)
        (5, 62.60)
        (6, 78.30)
        (7, 87.68)
        (8, 80.80)
        (9, 79.40)
        (10, 81.70)
        (11, 83.58)
        (12, 78.60)
        (13, 74.10)
        (14, 78.35)
        (15, 81.20)
        (16, 77.60)
        (17, 82.00)
        (18, 80.80)
        (19, 77.50)
        (20, 88.40)
        (21, 89.50)
        (22, 87.30)
        (23, 89.50)
    };

    \legend{Res-14 ,Lap-14}
    
    \end{axis}
    
    \draw [black] (current axis.south west) rectangle (current axis.north east);
    \end{tikzpicture}
    \end{adjustbox}
    \caption{Aspect extraction accuracy of models}
    \label{fig:aspect}
\end{figure}
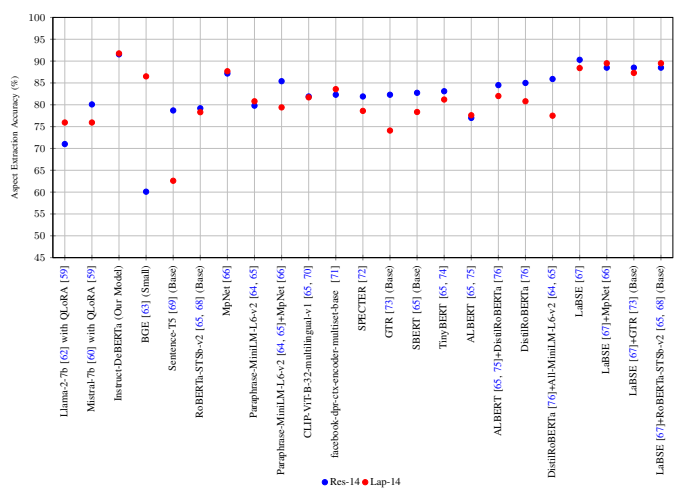

\section{Conclusion}

In this paper, we presented a comprehensive review and detailed experimental analysis of ABSA methodologies, focusing on the latest advancements in Transformer-based models. 


Our hybrid model, \texttt{Instruct-DeBERTa}, was designed to harness the specific advantages of two best performing models. \texttt{InstructABSA} is known for its accuracy in identifying and extracting relevant aspects from text, while \texttt{DeBERTa-V3-baseabsa-V1} excels in classifying the sentiment associated with these aspects. By integrating these models, we aimed to create a comprehensive tool that could perform both tasks with high precision and reliability.


Our model achieved the highest accuracy and F1 scores across multiple datasets, showcasing its ability to effectively handle diverse textual data and consistently deliver high-quality results. This performance can be attributed to the synergistic integration of \texttt{InstructABSA} and \texttt{DeBERTa-V3-baseabsa-V1}, which allows our hybrid model to maintain a delicate balance between precision in aspect extraction and accuracy in sentiment classification. 

In conclusion, our comprehensive review and experimental analysis highlight the significant advancements made possible by Transformer-based models in the field of ABSA. The development of \texttt{Instruct-DeBERTa} represents a notable contribution, offering a powerful and versatile solution for accurately extracting aspects and classifying sentiment in diverse textual data. The superior performance of our hybrid model sets a new benchmark for future research and applications in ABSA, underscoring the potential of integrating state-of-the-art models to enhance the effectiveness of sentiment analysis methodologies.

\balance

\bibliographystyle{IEEEtranN}

 \let\oldthebibliography\thebibliography
\renewcommand{\thebibliography}[1]{
   \oldthebibliography{#1}
   \small 
 }

\bibliography{References}

\begin{thebibliography}{76}
\providecommand{\natexlab}[1]{#1}
\providecommand{\url}[1]{#1}
\csname url@samestyle\endcsname
\providecommand{\newblock}{\relax}
\providecommand{\bibinfo}[2]{#2}
\providecommand{\BIBentrySTDinterwordspacing}{\spaceskip=0pt\relax}
\providecommand{\BIBentryALTinterwordstretchfactor}{4}
\providecommand{\BIBentryALTinterwordspacing}{\spaceskip=\fontdimen2\font plus
\BIBentryALTinterwordstretchfactor\fontdimen3\font minus \fontdimen4\font\relax}
\providecommand{\BIBforeignlanguage}[2]{{%
\expandafter\ifx\csname l@#1\endcsname\relax
\typeout{** WARNING: IEEEtranN.bst: No hyphenation pattern has been}%
\typeout{** loaded for the language `#1'. Using the pattern for}%
\typeout{** the default language instead.}%
\else
\language=\csname l@#1\endcsname
\fi
#2}}
\providecommand{\BIBdecl}{\relax}
\BIBdecl

\bibitem[Hochreiter and Schmidhuber(1997)]{hochreiter1997long}
S.~Hochreiter and J.~Schmidhuber, ``Long short-term memory,'' \emph{Neural computation}, vol.~9, no.~8, pp. 1735--1780, 1997.

\bibitem[Wang et~al.(2016)Wang, Huang, Zhu, and Zhao]{wang2016attention}
Y.~Wang, M.~Huang, X.~Zhu, and L.~Zhao, ``Attention-based lstm for aspect-level sentiment classification,'' in \emph{EMNLP}, 2016, pp. 606--615.

\bibitem[Li et~al.(2018)Li, Bing, Lam, and Shi]{li2018transformation}
X.~Li, L.~Bing, W.~Lam, and B.~Shi, ``Transformation networks for target-oriented sentiment classification,'' \emph{arXiv preprint arXiv:1805.01086}, 2018.

\bibitem[Zhang et~al.(2019)Zhang, Li, and Song]{zhang-etal-2019-aspect}
C.~Zhang, Q.~Li, and D.~Song, ``Aspect-based sentiment classification with aspect-specific graph convolutional networks,'' in \emph{EMNLP-IJCNLP}, 2019, pp. 4568--4578.

\bibitem[Li et~al.(2021)Li, Chen, Feng, Ma, Wang, and Hovy]{li2021dual}
R.~Li, H.~Chen, F.~Feng, Z.~Ma, X.~Wang, and E.~Hovy, ``Dual graph convolutional networks for aspect-based sentiment analysis,'' in \emph{ACL-IJCNLP}, 2021, pp. 6319--6329.

\bibitem[Zhong et~al.(2023)Zhong, Ding, Liu, Du, Jin, and Tao]{zhong2023KGAN}
Q.~Zhong, L.~Ding, J.~Liu, B.~Du, H.~Jin, and D.~Tao, ``Knowledge graph augmented network towards multiview representation learning for aspect-based sentiment analysis,'' \emph{IEEE Transactions on Knowledge and Data Engineering}, pp. 1--14, 2023.

\bibitem[Liu et~al.(2023)Liu, Zhong, Ding, Jin, Du, and Tao]{liu2023unified}
J.~Liu, Q.~Zhong, L.~Ding, H.~Jin, B.~Du, and D.~Tao, ``Unified instance and knowledge alignment pretraining for aspect-based sentiment analysis,'' \emph{IEEE/ACM transactions on audio, speech, and language processing}, vol.~31, pp. 2629--2642, 2023.

\bibitem[Devlin et~al.(2018)Devlin, Chang, Lee, and Toutanova]{devlin2018bert}
J.~Devlin, M.-W. Chang, K.~Lee, and K.~Toutanova, ``Bert: Pre-training of deep bidirectional transformers for language understanding,'' \emph{arXiv preprint arXiv:1810.04805}, 2018.

\bibitem[Zhao(2020)]{zhao2020bertdk}
e.~a. Zhao, ``Bert-dk: Incorporating domain knowledge into bert for knowledge-intensive tasks,'' \emph{arXiv preprint arXiv:2002.00000}, 2020.

\bibitem[Song et~al.(2019)Song, Wang, Jiang, Liu, and Rao]{song2019attentional}
Y.~Song, J.~Wang, T.~Jiang, Z.~Liu, and Y.~Rao, ``Attentional encoder network for targeted sentiment classification,'' \emph{arXiv preprint arXiv:1902.09314}, 2019.

\bibitem[Zhao et~al.(2020{\natexlab{a}})Zhao, Liu, Wang, and Zhang]{bert_mrc_2020}
X.~Zhao, Y.~Liu, J.~Wang, and L.~Zhang, ``Bert-mrc: A bert-based model for machine reading comprehension,'' \emph{Journal of Artificial Intelligence Research}, vol.~68, pp. 345--361, 2020.

\bibitem[Xu et~al.(2019)Xu, Liu, Shu, and Yu]{xu2019bert}
H.~Xu, B.~Liu, L.~Shu, and P.~S. Yu, ``Bert post-training for review reading comprehension and aspect-based sentiment analysis,'' \emph{arXiv preprint arXiv:1904.02232}, 2019.

\bibitem[Karimi et~al.(2021)Karimi, Rossi, and Prati]{karimi2021adversarial}
A.~Karimi, L.~Rossi, and A.~Prati, ``{Adversarial Training for Aspect-Based Sentiment Analysis with BERT},'' in \emph{ICPR}.\hskip 1em plus 0.5em minus 0.4em\relax IEEE, 2021, pp. 8797--8803.

\bibitem[Bai et~al.(2020)Bai, Liu, and Zhang]{bai2020investigating}
X.~Bai, P.~Liu, and Y.~Zhang, ``Investigating typed syntactic dependencies for targeted sentiment classification using graph attention neural network,'' \emph{IEEE/ACM Transactions on Audio, Speech, and Language Processing}, vol.~29, pp. 503--514, 2020.

\bibitem[Zhang et~al.(2023)Zhang, Zhu, Liu, Bao, Wu, Sun, and Xu]{zhang2023span}
M.~Zhang, Y.~Zhu, Z.~Liu, Z.~Bao, Y.~Wu, X.~Sun, and L.~Xu, ``Span-level aspect-based sentiment analysis via table filling,'' in \emph{ACl}, 2023, pp. 9273--9284.

\bibitem[Chen et~al.(2022)Chen, Teng, Wang, and Zhang]{chen2022discrete}
C.~Chen, Z.~Teng, Z.~Wang, and Y.~Zhang, ``Discrete opinion tree induction for aspect-based sentiment analysis,'' in \emph{ACl}, 2022, pp. 2051--2064.

\bibitem[Liu et~al.(2019)Liu, Ott, Goyal, Du, Joshi, Chen, Levy, Lewis, Zettlemoyer, and Stoyanov]{liu2019roberta}
Y.~Liu, M.~Ott, N.~Goyal, J.~Du, M.~Joshi, D.~Chen, O.~Levy, M.~Lewis, L.~Zettlemoyer, and V.~Stoyanov, ``Roberta: A robustly optimized bert pretraining approach,'' \emph{arXiv preprint arXiv:1907.11692}, 2019.

\bibitem[Wang et~al.(2021)Wang, Shen, Long, Zhou, and Chang]{wang2021eliminating}
B.~Wang, T.~Shen, G.~Long, T.~Zhou, and Y.~Chang, ``Eliminating sentiment bias for aspect-level sentiment classification with unsupervised opinion extraction,'' \emph{arXiv preprint arXiv:2109.02403}, 2021.

\bibitem[Dai et~al.(2021)Dai, Yan, Sun, Liu, and Qiu]{dai2021does}
J.~Dai, H.~Yan, T.~Sun, P.~Liu, and X.~Qiu, ``Does syntax matter? a strong baseline for aspect-based sentiment analysis with roberta,'' \emph{arXiv preprint arXiv:2104.04986}, 2021.

\bibitem[He et~al.(2020)He, Liu, Gao, and Chen]{he2020deberta}
P.~He, X.~Liu, J.~Gao, and W.~Chen, ``{DeBERTa: Decoding-enhanced BERT with Disentangled Attention},'' in \emph{ICLR}, 2020.

\bibitem[He et~al.(2021)He, Gao, and Chen]{he2021debertav3}
P.~He, J.~Gao, and W.~Chen, ``Debertav3: Improving deberta using electra-style pre-training with gradient-disentangled embedding sharing,'' \emph{arXiv preprint arXiv:2111.09543}, 2021.

\bibitem[Yang et~al.(2023)Yang, Zhang, and Li]{DBLP:conf/cikm/0008ZL23}
H.~Yang, C.~Zhang, and K.~Li, ``Pyabsa: {A} modularized framework for reproducible aspect-based sentiment analysis,'' in \emph{Proceedings of the 32nd {ACM} International Conference on Information and Knowledge Management, {CIKM} 2023, Birmingham, United Kingdom, October 21-25, 2023}, I.~Frommholz, F.~Hopfgartner, M.~Lee, M.~Oakes, M.~Lalmas, M.~Zhang, and R.~L.~T. Santos, Eds.\hskip 1em plus 0.5em minus 0.4em\relax {ACM}, 2023, pp. 5117--5122.

\bibitem[Marcacini and Silva(2021)]{marcacini2021aspect}
R.~Marcacini and E.~Silva, ``Aspect-based sentiment analysis using bert with disentangled attention,'' in \emph{LatinX in AI at ICML}, 2021, pp. 1--4.

\bibitem[Yang and Li(2024)]{yang2024modeling}
H.~Yang and K.~Li, ``{Modeling Aspect Sentiment Coherency via Local Sentiment Aggregation},'' in \emph{Findings of EACL}, 2024, pp. 182--195.

\bibitem[Yang et~al.(2021{\natexlab{a}})Yang, Zeng, Yang, Song, and Xu]{yang2021multi}
H.~Yang, B.~Zeng, J.~Yang, Y.~Song, and R.~Xu, ``A multi-task learning model for chinese-oriented aspect polarity classification and aspect term extraction,'' \emph{Neurocomputing}, vol. 419, pp. 344--356, 2021.

\bibitem[Scaria et~al.(2023)Scaria, Gupta, Goyal, Sawant, Mishra, and Baral]{scaria2023instructabsa}
K.~Scaria, H.~Gupta, S.~Goyal, S.~A. Sawant, S.~Mishra, and C.~Baral, ``Instructabsa: Instruction learning for aspect based sentiment analysis,'' \emph{arXiv preprint arXiv:2302.08624}, 2023.

\bibitem[Yang and Li(2021)]{yang2021improving}
H.~Yang and K.~Li, ``Modeling aspect sentiment coherency via local sentiment aggregation,'' \emph{arXiv preprint arXiv:2110.08604}, 2021.

\bibitem[Yan et~al.(2021)Yan, Dai, Qiu, Zhang, et~al.]{yan2021unified}
H.~Yan, J.~Dai, X.~Qiu, Z.~Zhang \emph{et~al.}, ``A unified generative framework for aspect-based sentiment analysis,'' \emph{arXiv preprint arXiv:2106.04300}, 2021.

\bibitem[Chen and Qian(2020)]{chen2020relation}
Z.~Chen and T.~Qian, ``Relation-aware collaborative learning for unified aspect-based sentiment analysis,'' in \emph{ACL}, 2020, pp. 3685--3694.

\bibitem[Hu et~al.(2019)Hu, Peng, Huang, Li, and Lv]{hu2019open}
M.~Hu, Y.~Peng, Z.~Huang, D.~Li, and Y.~Lv, ``Open-domain targeted sentiment analysis via span-based extraction and classification,'' \emph{arXiv preprint arXiv:1906.03820}, 2019.

\bibitem[Li et~al.(2019{\natexlab{a}})Li, Bing, Li, and Lam]{li2019unified}
X.~Li, L.~Bing, P.~Li, and W.~Lam, ``A unified model for opinion target extraction and target sentiment prediction,'' in \emph{AAAI}, vol.~33, no.~01, 2019, pp. 6714--6721.

\bibitem[Li et~al.(2019{\natexlab{b}})Li, Bing, Zhang, and Lam]{li2019exploiting}
X.~Li, L.~Bing, W.~Zhang, and W.~Lam, ``Exploiting bert for end-to-end aspect-based sentiment analysis,'' \emph{arXiv preprint arXiv:1910.00883}, 2019.

\bibitem[Luo et~al.(2019)Luo, Li, Liu, and Zhang]{luo2019doer}
H.~Luo, T.~Li, B.~Liu, and J.~Zhang, ``Doer: Dual cross-shared rnn for aspect term-polarity co-extraction,'' \emph{arXiv preprint arXiv:1906.01794}, 2019.

\bibitem[He et~al.(2019)He, Lee, Ng, and Dahlmeier]{he2019interactive}
R.~He, W.~S. Lee, H.~T. Ng, and D.~Dahlmeier, ``An interactive multi-task learning network for end-to-end aspect-based sentiment analysis,'' \emph{arXiv preprint arXiv:1906.06906}, 2019.

\bibitem[Pontiki et~al.(2014)Pontiki, Galanis, Pavlopoulos, Papageorgiou, Androutsopoulos, and Manandhar]{pontiki-etal-2014-semeval}
M.~Pontiki, D.~Galanis, J.~Pavlopoulos, H.~Papageorgiou, I.~Androutsopoulos, and S.~Manandhar, ``{S}em{E}val-2014 task 4: Aspect based sentiment analysis,'' in \emph{{S}em{E}val 2014}, 2014, pp. 27--35.

\bibitem[Luo et~al.(2020)Luo, Ji, Li, Duan, and Jiang]{luo2020grace}
H.~Luo, L.~Ji, T.~Li, N.~Duan, and D.~Jiang, ``Grace: Gradient harmonized and cascaded labeling for aspect-based sentiment analysis,'' \emph{arXiv preprint arXiv:2009.10557}, 2020.

\bibitem[Xu et~al.(2020)Xu, Bing, Lu, and Huang]{xu2020aspect}
L.~Xu, L.~Bing, W.~Lu, and F.~Huang, ``Aspect sentiment classification with aspect-specific opinion spans,'' in \emph{EMNLP}, 2020, pp. 3561--3567.

\bibitem[Xing and Tsang(2022)]{xing2022understand}
B.~Xing and I.~W. Tsang, ``Understand me, if you refer to aspect knowledge: Knowledge-aware gated recurrent memory network,'' \emph{IEEE Transactions on Emerging Topics in Computational Intelligence}, vol.~6, no.~5, pp. 1092--1102, 2022.

\bibitem[Ma et~al.(2017)Ma, Li, Zhang, and Wang]{ma2017interactive}
D.~Ma, S.~Li, X.~Zhang, and H.~Wang, ``Interactive attention networks for aspect-level sentiment classification,'' \emph{arXiv preprint arXiv:1709.00893}, 2017.

\bibitem[Yang et~al.(2021{\natexlab{b}})Yang, Zeng, Xu, and Wang]{YangZMT21}
H.~Yang, B.~Zeng, M.~Xu, and T.~Wang, ``Back to reality: Leveraging pattern-driven modeling to enable affordable sentiment dependency learning,'' \emph{arXiv preprint arXiv:2110.08604}, 2021.

\bibitem[Karimi et~al.(2020)Karimi, Rossi, and Prati]{karimi2020improving}
A.~Karimi, L.~Rossi, and A.~Prati, ``Improving bert performance for aspect-based sentiment analysis,'' \emph{arXiv preprint arXiv:2010.11731}, 2020.

\bibitem[Zhou et~al.(2020)Zhou, Huang, Hu, and He]{zhou2020sk}
J.~Zhou, J.~X. Huang, Q.~V. Hu, and L.~He, ``Sk-gcn: Modeling syntax and knowledge via graph convolutional network for aspect-level sentiment classification,'' \emph{Knowledge-Based Systems}, vol. 205, p. 106292, 2020.

\bibitem[Zhao et~al.(2020{\natexlab{b}})Zhao, Hou, and Wu]{zhao2020modeling}
P.~Zhao, L.~Hou, and O.~Wu, ``Modeling sentiment dependencies with graph convolutional networks for aspect-level sentiment classification,'' \emph{Knowledge-Based Systems}, vol. 193, p. 105443, 2020.

\bibitem[Tang et~al.(2020)Tang, Ji, Li, and Zhou]{tang2020dependency}
H.~Tang, D.~Ji, C.~Li, and Q.~Zhou, ``Dependency graph enhanced dual-transformer structure for aspect-based sentiment classification,'' in \emph{Proceedings of the 58th annual meeting of the ACL}, 2020, pp. 6578--6588.

\bibitem[Rietzler et~al.(2019)Rietzler, Stabinger, Opitz, and Engl]{rietzler2019adapt}
A.~Rietzler, S.~Stabinger, P.~Opitz, and S.~Engl, ``Adapt or get left behind: Domain adaptation through bert language model finetuning for aspect-target sentiment classification,'' \emph{arXiv preprint arXiv:1908.11860}, 2019.

\bibitem[Mao et~al.(2021)Mao, Shen, Yu, and Cai]{mao2021joint}
Y.~Mao, Y.~Shen, C.~Yu, and L.~Cai, ``A joint training dual-mrc framework for aspect based sentiment analysis,'' in \emph{Proceedings of the AAAI conference on artificial intelligence}, vol.~35, no.~15, 2021, pp. 13\,543--13\,551.

\bibitem[Zhang et~al.(2022{\natexlab{a}})Zhang, Zhang, Wu, and Zhao]{zhang2022towards}
Y.~Zhang, M.~Zhang, S.~Wu, and J.~Zhao, ``Towards unifying the label space for aspect- and sentence-based sentiment analysis,'' \emph{arXiv preprint arXiv:2203.07090}, 2022.

\bibitem[Zhang et~al.(2022{\natexlab{b}})Zhang, Zhou, and Wang]{zhang2022ssegcn}
Z.~Zhang, Z.~Zhou, and Y.~Wang, ``Ssegcn: Syntactic and semantic enhanced graph convolutional network for aspect-based sentiment analysis,'' in \emph{NAACL}, 2022, pp. 4916--4925.

\bibitem[Liang et~al.(2022)Liang, Su, Gui, Cambria, and Xu]{liang2022aspect}
B.~Liang, H.~Su, L.~Gui, E.~Cambria, and R.~Xu, ``Aspect-based sentiment analysis via affective knowledge enhanced graph convolutional networks,'' \emph{Knowledge-Based Systems}, vol. 235, p. 107643, 2022.

\bibitem[Xu et~al.(2021)Xu, Chia, and Bing]{Xu2021SpanLevel}
L.~Xu, Y.~K. Chia, and L.~Bing, ``Learning span-level interactions for aspect sentiment triplet extraction,'' \emph{ACL}, 2021.

\bibitem[Xue and Li(2018)]{xue2018aspect}
W.~Xue and T.~Li, ``Aspect based sentiment analysis with gated convolutional networks,'' in \emph{ACL}, 2018, pp. 2514--2523.

\bibitem[Chen and Qian(2019)]{chen2019transfer}
Z.~Chen and T.~Qian, ``Transfer capsule network for aspect level sentiment classification,'' in \emph{ACL}, 2019, pp. 547--556.

\bibitem[Wang et~al.(2018)Wang, Mazumder, Liu, Zhou, and Chang]{wang2018target}
S.~Wang, S.~Mazumder, B.~Liu, M.~Zhou, and Y.~Chang, ``Target-sensitive memory networks for aspect sentiment classification,'' in \emph{ACL}, 2018.

\bibitem[Tang et~al.(2019)Tang, Lu, Su, Ge, Song, Sun, and Luo]{tang2019progressive}
J.~Tang, Z.~Lu, J.~Su, Y.~Ge, L.~Song, L.~Sun, and J.~Luo, ``Progressive self-supervised attention learning for aspect-level sentiment analysis,'' \emph{arXiv preprint arXiv:1906.01213}, 2019.

\bibitem[Nguyen and Le~Nguyen(2018)]{nguyen2018effective}
H.~T. Nguyen and M.~Le~Nguyen, ``Effective attention modeling for aspect-level sentiment classification,'' in \emph{2018 10th International conference on knowledge and systems engineering (KSE)}.\hskip 1em plus 0.5em minus 0.4em\relax IEEE, 2018, pp. 25--30.

\bibitem[Roumeliotis et~al.(2024)Roumeliotis, Tselikas, and Nasiopoulos]{roumeliotis2024llms}
K.~I. Roumeliotis, N.~D. Tselikas, and D.~K. Nasiopoulos, ``Llms in e-commerce: a comparative analysis of gpt and llama models in product review evaluation,'' \emph{Natural Language Processing Journal}, vol.~6, p. 100056, 2024.

\bibitem[Peng et~al.(2023)Peng, Li, He, Galley, and Gao]{peng2023instruction}
B.~Peng, C.~Li, P.~He, M.~Galley, and J.~Gao, ``Instruction tuning with gpt-4,'' \emph{arXiv preprint arXiv:2304.03277}, 2023.

\bibitem[Hu et~al.(2021)Hu, Wallis, Allen-Zhu, Li, Wang, Wang, Chen, et~al.]{hu2021lora}
E.~J. Hu, P.~Wallis, Z.~Allen-Zhu, Y.~Li, S.~Wang, L.~Wang, W.~Chen \emph{et~al.}, ``{LoRA: Low-Rank Adaptation of Large Language Models},'' in \emph{ICLR}, 2021.

\bibitem[Dettmers et~al.(2023)Dettmers, Pagnoni, Holtzman, and Zettlemoyer]{dettmers2023qlora}
T.~Dettmers, A.~Pagnoni, A.~Holtzman, and L.~Zettlemoyer, ``{QLoRA: Efficient Finetuning of Quantized LLMs},'' in \emph{NeurIPS}, vol.~36, 2023.

\bibitem[Jiang et~al.(2023)Jiang, Sablayrolles, Mensch, Bamford, Chaplot, Casas, Bressand, Lengyel, Lample, Saulnier, et~al.]{jiang2023mistral}
A.~Q. Jiang, A.~Sablayrolles, A.~Mensch, C.~Bamford, D.~S. Chaplot, D.~d.~l. Casas, F.~Bressand, G.~Lengyel, G.~Lample, L.~Saulnier \emph{et~al.}, ``Mistral 7b,'' \emph{arXiv preprint arXiv:2310.06825}, 2023.

\bibitem[Tunstall et~al.(2022)Tunstall, Reimers, Jo, Bates, Korat, Wasserblat, and Pereg]{tunstall2022efficient}
L.~Tunstall, N.~Reimers, U.~E.~S. Jo, L.~Bates, D.~Korat, M.~Wasserblat, and O.~Pereg, ``Efficient few-shot learning without prompts,'' \emph{arXiv preprint arXiv:2209.11055}, 2022.

\bibitem[Touvron et~al.(2023)Touvron, Martin, Stone, Albert, Almahairi, Babaei, Bashlykov, Batra, Bhargava, Bhosale, et~al.]{touvron2023llama}
H.~Touvron, L.~Martin, K.~Stone, P.~Albert, A.~Almahairi, Y.~Babaei, N.~Bashlykov, S.~Batra, P.~Bhargava, S.~Bhosale \emph{et~al.}, ``{Llama 2: Open Foundation and Fine-Tuned Chat Models},'' \emph{arXiv preprint arXiv:2307.09288}, 2023.

\bibitem[Xiao et~al.(2023)Xiao, Liu, Zhang, and Muennighof]{xiao2023c}
S.~Xiao, Z.~Liu, P.~Zhang, and N.~Muennighof, ``{C-Pack: Packaged Resources To Advance General Chinese Embedding},'' \emph{arXiv preprint arXiv:2309.07597}, 2023.

\bibitem[Wang et~al.(2020)Wang, Wei, Dong, Bao, Yang, and Zhou]{wang2020minilm}
W.~Wang, F.~Wei, L.~Dong, H.~Bao, N.~Yang, and M.~Zhou, ``{MiniLM: Deep Self-Attention Distillation for Task-Agnostic Compression of Pre-Trained Transformers},'' \emph{NeurIPS}, vol.~33, pp. 5776--5788, 2020.

\bibitem[Reimers and Gurevych(2019)]{reimers2019sentence}
N.~Reimers and I.~Gurevych, ``{Sentence-BERT: Sentence Embeddings using Siamese BERT-Networks},'' in \emph{EMNLP-IJCNLP}, 2019, pp. 3982--3992.

\bibitem[Song et~al.(2020)Song, Tan, Qin, Lu, and Liu]{song2020mpnet}
K.~Song, X.~Tan, T.~Qin, J.~Lu, and T.-Y. Liu, ``{MPNet: Masked and Permuted Pre-training for Language Understanding},'' \emph{NeurIPS}, vol.~33, pp. 16\,857--16\,867, 2020.

\bibitem[Feng et~al.(2022)Feng, Yang, Cer, Arivazhagan, and Wang]{feng2022language}
F.~Feng, Y.~Yang, D.~Cer, N.~Arivazhagan, and W.~Wang, ``{Language-agnostic BERT Sentence Embedding},'' in \emph{ACL}, 2022, pp. 878--891.

\bibitem[Cer et~al.(2017)Cer, Diab, Agirre, Lopez-Gazpio, and Specia]{cer2017semeval}
D.~Cer, M.~Diab, E.~Agirre, I.~Lopez-Gazpio, and L.~Specia, ``{SemEval-2017 Task 1: Semantic Textual Similarity Multilingual and Crosslingual Focused Evaluation},'' in \emph{SemEval}, 2017.

\bibitem[Ni et~al.(2022{\natexlab{a}})Ni, Abrego, Constant, Ma, Hall, Cer, and Yang]{ni2022sentence}
J.~Ni, G.~H. Abrego, N.~Constant, J.~Ma, K.~Hall, D.~Cer, and Y.~Yang, ``{Sentence-T5: Scalable Sentence Encoders from Pre-trained Text-to-Text Models},'' in \emph{Findings of ACL}, 2022, pp. 1864--1874.

\bibitem[Radford et~al.(2021)Radford, Kim, Hallacy, Ramesh, Goh, Agarwal, Sastry, Askell, Mishkin, Clark, et~al.]{radford2021learning}
A.~Radford, J.~W. Kim, C.~Hallacy, A.~Ramesh, G.~Goh, S.~Agarwal, G.~Sastry, A.~Askell, P.~Mishkin, J.~Clark \emph{et~al.}, ``{Learning transferable visual models from natural language supervision},'' in \emph{ICML}.\hskip 1em plus 0.5em minus 0.4em\relax PMLR, 2021, pp. 8748--8763.

\bibitem[Karpukhin et~al.(2020)Karpukhin, Oguz, Min, Lewis, Wu, Edunov, Chen, and Yih]{karpukhin-etal-2020-dense}
V.~Karpukhin, B.~Oguz, S.~Min, P.~Lewis, L.~Wu, S.~Edunov, D.~Chen, and W.-t. Yih, ``Dense passage retrieval for open-domain question answering,'' in \emph{Proceedings of the 2020 Conference on EMNLP}.\hskip 1em plus 0.5em minus 0.4em\relax Online: ACL, Nov. 2020, pp. 6769--6781.

\bibitem[Cohan et~al.(2020)Cohan, Feldman, Beltagy, Downey, and Weld]{specter2020cohan}
A.~Cohan, S.~Feldman, I.~Beltagy, D.~Downey, and D.~S. Weld, ``{SPECTER: Document-level Representation Learning using Citation-informed Transformers},'' in \emph{ACL}, 2020.

\bibitem[Ni et~al.(2022{\natexlab{b}})Ni, Qu, Lu, Dai, Abrego, Ma, Zhao, Luan, Hall, Chang, et~al.]{ni2022large}
J.~Ni, C.~Qu, J.~Lu, Z.~Dai, G.~H. Abrego, J.~Ma, V.~Zhao, Y.~Luan, K.~Hall, M.-W. Chang \emph{et~al.}, ``{Large Dual Encoders Are Generalizable Retrievers},'' in \emph{EMNLP}, 2022, pp. 9844--9855.

\bibitem[Jiao et~al.(2020)Jiao, Yin, Shang, Jiang, Chen, Li, Wang, and Liu]{jiao2020tinybert}
X.~Jiao, Y.~Yin, L.~Shang, X.~Jiang, X.~Chen, L.~Li, F.~Wang, and Q.~Liu, ``{TinyBERT: Distilling BERT for Natural Language Understanding},'' in \emph{Findings of EMNLP}, 2020, pp. 4163--4174.

\bibitem[Lan et~al.(2019)Lan, Chen, Goodman, Gimpel, Sharma, and Soricut]{lan2019albert}
Z.~Lan, M.~Chen, S.~Goodman, K.~Gimpel, P.~Sharma, and R.~Soricut, ``{ALBERT: A Lite BERT for Self-supervised Learning of Language Representations},'' \emph{arXiv preprint arXiv:1909.11942}, 2019.

\bibitem[Sanh et~al.(2019)Sanh, Debut, Chaumond, and Wolf]{Sanh2019DistilBERTAD}
V.~Sanh, L.~Debut, J.~Chaumond, and T.~Wolf, ``{DistilBERT, a distilled version of BERT: smaller, faster, cheaper and lighter},'' \emph{ArXiv}, vol. abs/1910.01108, 2019.

\end{thebibliography}
\end{document}